%% file: sample-acmtog-SIGGRAPH-submission.tex
\documentclass[sigconf]{acmart}

\usepackage{multirow}
\usepackage{makecell}

\usepackage{subcaption}
\usepackage{stfloats}

\usepackage{cleveref}
\usepackage{siunitx}

\setcopyright{none}
\renewcommand\footnotetextcopyrightpermission[1]{} 
\acmConference[]{}{}

\begin{document}
\title{AONeuS: A Neural Rendering Framework for Acoustic-Optical Sensor Fusion}

\author{Mohamad Qadri}
\email{mqadri@andrew.cmu.edu}
\authornote{Both authors contributed equally to the paper.}
\affiliation{%
 \institution{Carnegie Mellon University }
 \city{Pittsburgh}
 \country{USA}
 }

\author{Kevin Zhang}
\email{kzhang24@umd.edu}
\authornotemark[1]
\affiliation{%
 \institution{University of Maryland}
 \city{College Park}
 \country{USA}
}

\author{Akshay Hinduja}
\email{ahinduja@andrew.cmu.edu}
\affiliation{%
 \institution{Carnegie Mellon University }
 \city{Pittsburgh}
 \country{USA}
 }

\author{Michael Kaess}
\email{kaess@cmu.edu}
\affiliation{%
 \institution{Carnegie Mellon University }
 \city{Pittsburgh}
 \country{USA}
 }

\author{Adithya Pediredla}
\email{Adithya.K.Pediredla@dartmouth.edu}
\affiliation{%
 \institution{Dartmouth College}
 \city{Hanover}
 \country{USA}
}

\author{Christopher A.~Metzler}
\email{metzler@umd.edu}
\authornote{Corresponding author.}
\affiliation{%
 \institution{University of Maryland}
 \city{College Park}
 \country{USA}
}

\input{sec/0_abstract}
\begin{teaserfigure}
  \centering
  \Description{A teaser figure for our paper showing that our method, AONeuS, performs better than the baselines, NeuSIS and NeuS.}
   \includegraphics[width=.95\linewidth]{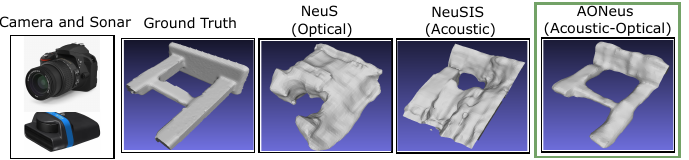}
   \caption{\textbf{AONeuS Experimental Results}: Under the restricted baseline operating conditions commonly encountered in underwater construction and navigation, camera-only reconstruction techniques (NeuS~\cite{wang2021neus}) and sonar-only reconstruction techniques (NeuSIS~\cite{qadri2023neural}) struggle to accurately recover 3D surface geometry. 
   This is due to the highly underdetermined nature of their respective measurement processes; cameras lack depth information, and imaging sonars do not capture elevation information. 
   We have developed a multimodal acoustic-optical neural surfaces reconstruction framework (AONeuS) that effectively combines data from these complementary modalities. }
   \label{fig:teaser}
\end{teaserfigure}

\ccsdesc[500]{Computing methodologies~3D imaging}

\keywords{3D reconstruction, implicit neural representations, multimodal sensing,
robotics, imaging sonar, underwater imaging, signed distance functions, neural rendering, inverse rendering, sensor fusion, underwater sensing}

\maketitle
\renewcommand{\shortauthors}{Qadri et al.}

\input{sec/1_intro}

\input{sec/2_related_work}

\input{sec/3_method}

\input{sec/4_experimental_results}

\input{sec/5_discussion}

\input{sec/7_ack}

\newpage
\bibliographystyle{ACM-Reference-Format}
\bibliography{sample-bibliography}
\input{sec/6_figure_only}
\clearpage
\appendix 
\section{Supplement}

\input{sec/X_suppl}

\end{document}

%% file: sec/0_abstract.tex
\begin{abstract}
Underwater perception and 3D surface reconstruction are challenging problems with broad applications in construction, security, marine archaeology, and environmental monitoring. Treacherous operating conditions, fragile surroundings, and limited navigation control often dictate that submersibles restrict their range of motion and, thus, the baseline over which they can capture measurements. In the context of 3D scene reconstruction, it is well-known that smaller baselines make reconstruction more challenging.  Our work develops a physics-based multimodal acoustic-optical neural surface reconstruction framework (AONeuS) capable of effectively integrating high-resolution RGB measurements with low-resolution depth-resolved imaging sonar measurements. By fusing these complementary modalities, our framework can reconstruct accurate high-resolution 3D surfaces from measurements captured over heavily-restricted baselines. Through extensive simulations and in-lab experiments, we demonstrate that AONeuS dramatically outperforms recent RGB-only and sonar-only inverse-differentiable-rendering--based surface reconstruction methods. 
\end{abstract}

%% file: sec/1_intro.tex
\section{Introduction}
\label{sec:intro}

The 3D reconstruction of underwater environments is an important problem with applications in myriad fields, including underwater construction, marine ecology, archaeology, mapping, inspection, and surveillance \cite{albiez2015flatfish, lin2023conditional, wang2019underwater, negahdaripour2018application}. The underwater robots applied to this task are typically equipped with both imaging sonars (i.e., acoustic cameras) and optical cameras~\cite{lensgraf2021droplet,liu2023deep}. These sensors capture complementary information about their operating environments. 

Forward-look imaging sonars consist of a uniform linear array of transducers which, through beamforming, recover both range and azimuth information (but not elevation). (3D imaging sonars which record both azimuth and elevation also exist, but can be prohibitively expensive.)  Unlike light-based sensors, imaging sonars are highly robust to scattering and low-light conditions. 
Unfortunately, imaging sonars generally have poor spatial resolution; sonar images of an object of interest often appear textureless and hard to recognize; and imaging sonar measurements can suffer from complex artifacts caused by multipath reflections and the variable speed of sound passing through inhomogeneous water~\cite{tinh2021new}.

By contrast, optical cameras have high spatial resolution and can resolve object appearance in great detail. However, in turbid water light scattering and absorption can severely restrict the range and contrast of optical cameras~\cite{jaffe2014underwater}. Moreover, to recover depth information passive optical sensors rely on large displacements/baselines between measurements. In constrained operating environments, such measurements are often inaccessible.

By leveraging the complementary strengths and weaknesses of cameras and imaging sonars, acoustic-optical sensor fusion promises to enable robust and high-resolution underwater perception and scene reconstruction~\cite{ferreira2016underwater,menna2018state}. 
Existing contour matching based acoustic-optical reconstruction methods can already reconstruct accurate high-resolution 3D surfaces~\cite{babaee20153}. 
Unfortunately, these methods require a 360-degree view of the scene and are inapplicable in the small-baseline operating conditions prevalent in real-world unmanned underwater vehicle operation.
Alternatively, one can reconstruct the scene from optical and acoustic measurements independently and then fuse the result~\cite{kim3DReconstructionUnderwater2019}. 
However, this simple approach provides limited benefits over camera-only surface reconstruction. 

In this paper, we develop an inverse-differentiable-rendering--based approach to acoustic-optical sensor fusion that can form dense 3D surface reconstructions from camera and sonar measurements captured across a small baseline. Our work consists of four key contributions.
\begin{itemize}
    \item We develop a physics-based multimodal acoustic-optical neural surface framework which simultaneously integrates RGB and imaging sonar measurements. Our approach extends the neural surfaces 3D reconstruction framework~\cite{wang2021neus} by combining a unified representation of the scene geometry with modality-specific (acoustic and optical) representations of appearance.
    \item We conduct experiments on both synthetic and experimentally-captured datasets and demonstrate our method can effectively reconstruct high-fidelity surface geometry from noisy measurements captured over limited baselines. 
    \item We theoretically support our strong empirical performance by analyzing the conditioning of the acoustic-optical forward model. We show that the forward process associated with triangulating a point in 3D from acoustic-optical measurements is better conditioned and easier to invert the unimodal forward models.
    \item We release a public dataset and open-source implementation of our method. 
\end{itemize}

%% file: sec/2_related_work.tex
\section{Related Work}

\label{sec:related_work}
\paragraph{Camera Imaging}
Myriad works have investigated the use of (optical) cameras for underwater 3D imaging. \citet{johnson2010generation} proposed a feature-based stereo-optical SLAM system for building 3D models. \citet{iscar2017multi} presented a comprehensive evaluation of different monocular and stereo software and hardware systems targeting underwater imaging. However, these techniques, which do not use sonar, have limited 
depth resolution in small baseline scenarios. \citet{roznere20233} proposed a multi-view photometric stereo method for non-stationary underwater robots 3D reconstruction that integrates ORB-SLAM \cite{mur2015orb} with traditional photometric stereo. However, this technique requires active illumination and is sensitive to backscattering in turbid waters.
\paragraph{Sonar Imaging}
3D reconstruction from sonar imagery is an important and widely studied problem. Over the last decade a variety of 3D reconstruction methods have been proposed based on space carving~\cite{aykin20153, aykin2016three}, classical point-cloud processing algorithms~\cite{teixeira2016underwater,westman2020theory}, generative modeling~\cite{aykin20153, aykin2016modeling, negahdaripour2017refining,westman2019wide}, convex optimization~\cite{westman2020volumetric}, graph-based processing~\cite{wang20183d,wang2019three}, and supervised machine learning~\cite{debortoli2019elevatenet, wang2021elevation,arnold2022spatial}. 

Last year, two research groups employed neural rendering to enable breakthrough 3D sonar imaging performance.~\citet{qadri2023neural} developed a Neural Implicit Surface Reconstruction Using Imaging Sonar (NeuSIS) method which forms high-fidelity 3D surface reconstructions from forward imaging sonar measurements by combining neural surface representations with a novel acoustic differentiable volumetric renderer. 
Similarly,~\citet{reedNeuralVolumetricReconstruction2023} employed neural rendering to recover 3D volumes from synthetic aperture sonar measurements.
The former method relies upon a large number of sonar images captured over a large baseline while the latter applies to synthetic aperture sonar, not forward imaging sonar, and relies on access to raw time-based sonar measurements. 
These methods represent the state-of-the-art in 3D surface reconstruction with sonar.

To date, no method has effectively recovered a dense 3D scene from 2D sonar images captured over a limited baseline. Without additional constrains, e.g.,~optical measurements, or strong priors the reconstruction problem is hopelessly underdetermined.

\paragraph{Neural Rendering}
In their breakthrough neural radiance fields (NeRF) paper,~\citet{mildenhall2020nerf} combined neural signal representations with differentiable volume rendering to perform novel view synthesis.
The underlying differentiable volume rendering concept has since been extended to represent and recover scene geometry. 
The Implicit Differentiable Renderer (IDR) approach, introduced in ~\cite{yariv2020multiview}, represents geometry as the zero-level set of a neural network and uses differentiable surface rendering to fit the parameters of a neural network. IDR requires object masks for supervision. 
Later methods, like Neural Surfaces (NeuS)~\cite{wang2021neus}, Unified Surfaces (UNISURF)~\cite{Oechsle2021ICCV}, and Volume Signed Distance Functions (VolSDF)~\cite{yariv2021volume} combine an implicit surface representation with differentiable volume rendering to recover 3D geometry from images without the need for object masks. 
Recent work has sought to reduce the number of training images required~\cite{long2022sparseneus} and to accelerate rendering to enable real-time applications~\cite{yariv2023bakedsdf}.
The inverse-differentiable-rendering framework has also been extended to handle measurements from a diverse range of sensors.
Cross-spectral radiance fields (X-NeRF) were proposed in~\cite{poggi2022xnerf} to model multispectral, infrared, and RGB images. 
Transient neural radiance fields were proposed in~\cite{malik2023transient} to model the measurements from a single-photon lidar. 
Time-of-flight radiance fields were proposed in ~\cite{attal2021torf} to model the measurements from a continuous wave time-of-flight sensor. 
Polarization-aided decomposition of radiance, or PANDORA, was proposed in~\cite{Dave2022} to model polarimetric measurements of light.  
Radar neural radiance fields (RaNeRF) were proposed in~\cite{10190736} to model inverse synthetic aperture radar measurements. 

Recent research on neural fields has also addressed the small baseline 3D reconstruction problem we tackle in our paper, but leveraging natural small hand motions to improve depth reconstruction instead of using multiple imaging modalities ~\cite{chugunov2022implicit,chugunov2023shakes,chugunov2024neural}.

Several recent works have modeled light scattering within the neural rendering framework to improve reconstructions through water~\cite{sethuraman2023waternerf,levy2023seathru}, haze~\cite{Chen2024dehazenerf}, and fog~\cite{ramazzina2023scatternerf}.

\paragraph{Multimodal Imaging}
To overcome the disadvantages inherent to using a single sensing modality, numerous multimodal sensing algorithms have been developed~\cite{5457430,Lindell:2018:3D,nishimura2020disambiguating,bijelic2020seeing}. Most related to our work,~\citet{babaee20153} reconstruct 3D objects from RGB and sonar imagery by matching occluding contours across RGB images and imaging sonar measurements, performing stereo matching, and interpolating the curves in 3D space. Unfortunately, this method is inapplicable to the small-baseline setting; it fundamentally requires 360-degree views of the scene.
~\citet{cardaillac2023camera} similarly use a camera and an imaging sonar for 3D reconstruction by matching features between the acoustic and optical measurements. However, this matching can be frail and prone to errors.
~\citet{kim3DReconstructionUnderwater2019} reconstructs a scene from optical and acoustic measurements independently using classical methods like COLMAP~\cite{schonberger2016structure} and then fuses the result. As we demonstrate in~\cref{ssec:ExpResults}, this approach provides limited benefits over a purely optical approach and is not competitive with state-of-the-art neural rendering based methods.

Outside of sonar, several neural-rendering based approaches to sensor fusion have recently been developed. 
In Multimodal Neural Radiance Field,~\citet{10160388} use neural rendering to combine RGB, thermal, and point cloud data. 
Similarly,~\cite{10.1145/3610548.3618172} use neural rendering to combine multispectral measurements of different polarizations and~\citet{10081483} fuse sparse lidar and RGB measurements to build 3D occupancy grid of unbounded scenes. 

To our knowledge, ours is the first work to perform acoustic-optical sensor fusion with neural rendering.

%% file: sec/3_method.tex
\section{Background}
\begin{figure}[h!] 
  \centering
  \Description{A figure that graphically shows the acoustic and image formation models, comparing and contrasting them.}
   \includegraphics[width=.95\linewidth]{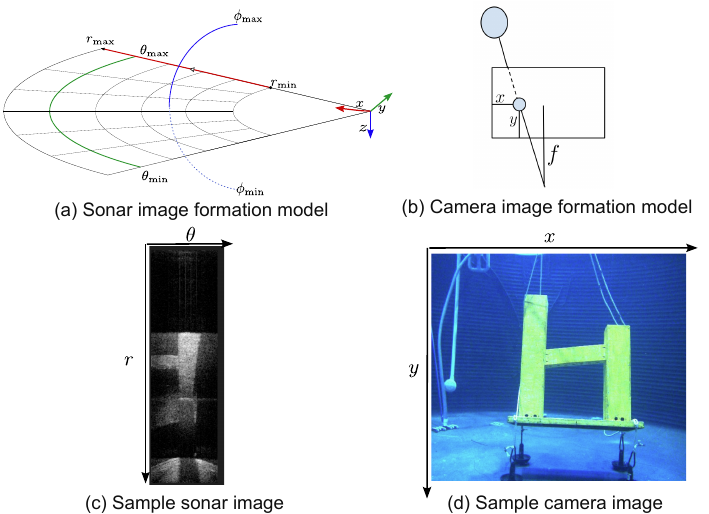}
   \caption{\textbf{Acoustic-Optical Measurement Processes.} (a) RGB measurement process and example measurement. Pixels along a common ray passing through the camera center map to the same image pixel on the image plane. (b) Sonar measurement process and example measurement. In a sonar image, the azimuth $\theta$ and range $r$ of the imaged object are resolved. However, the elevation information $\phi$ is lost; all objects located along the elevation arc (in blue) map to the same pixel.}
   \label{fig:Measurements}
   \vspace{-15pt}
\end{figure}
\subsection{Imaging Sonars}
Imaging sonars are active sensors that emit acoustic pulses and measure the intensity of the reflected wave. They produce a 2D acoustic image in which the range and azimuth of the imaged object are resolved. However, the object's elevation remains ambiguous. I.e., the reflecting object can be located anywhere on the elevation arc (\cref{fig:Measurements}) and the intensity of a pixel in a sonar image is proportional to the cumulative reflected acoustic energy from all reflecting points along the elevation arc.

\subsection{Image Formation Model of an Imaging Sonar}
Similar to \cite{qadri2023neural}, we use the following sonar image formation model: 
\begin{align}
& \!\! \!\! I^{\text{son}}(r_i, \theta_i)\!=\!\! \int_{\phi_\text{min}}^{\phi_\text{max}} \! \int_{r_i-\epsilon}^{r_i+\epsilon} \frac{\textit{E}_e}{r} T(r, \theta_i, \phi) \sigma(r, \theta_i, \phi) \text{d}r\text{d}\phi, 
\label{renderinglossSonar}
\end{align}
where $\phi_\text{min}, \phi_\text{max}$ are the minimum and maximum elevation angles,  $E_e$ is the acoustic energy emitted by the sonar. $T=e^{-\int_0^{r_i} \sigma(r', \theta_i, \phi_i ) \text{d}r'}$ is the transmittance term, and $\sigma$ is the particle density. (See \cite{qadri2023neural} for more details.)
\subsection{Image Formation Model of an Optical Camera}
We adopt the optical camera image formation model proposed by \cite{wang2021neus} where a pixel intensity at $(x,y)$ is approximated by: 
\begin{align}
    I^{\text{cam}}(x,y) = \int_{0}^{\infty} T(t) \sigma(t) c(\mathbf{p}(t), \mathbf{v}) dt, 
    \label{renderinglossCam}
\end{align}
where the integral is over the ray starting at the camera center and passing through pixel $(x,y)$. $T,\sigma$ are the transmittance and density values at point $p(t)$, and $c(\mathbf{p}(t), \mathbf{v})$ is the color of a point viewed from direction $\mathbf{v}$.

\section{Problem Statement}
\begin{figure}[h!]
  \centering
  \Description{A figure that shows how stereo with RGB measurements has ambiguities in the depth direction, and stereo with imaging sonar measurements has ambiguities in the horizontal direction, so RGB and imaging sonar stereo having orthogonal ambiguities and are thus complementary.}
   \includegraphics[width=.95\linewidth]{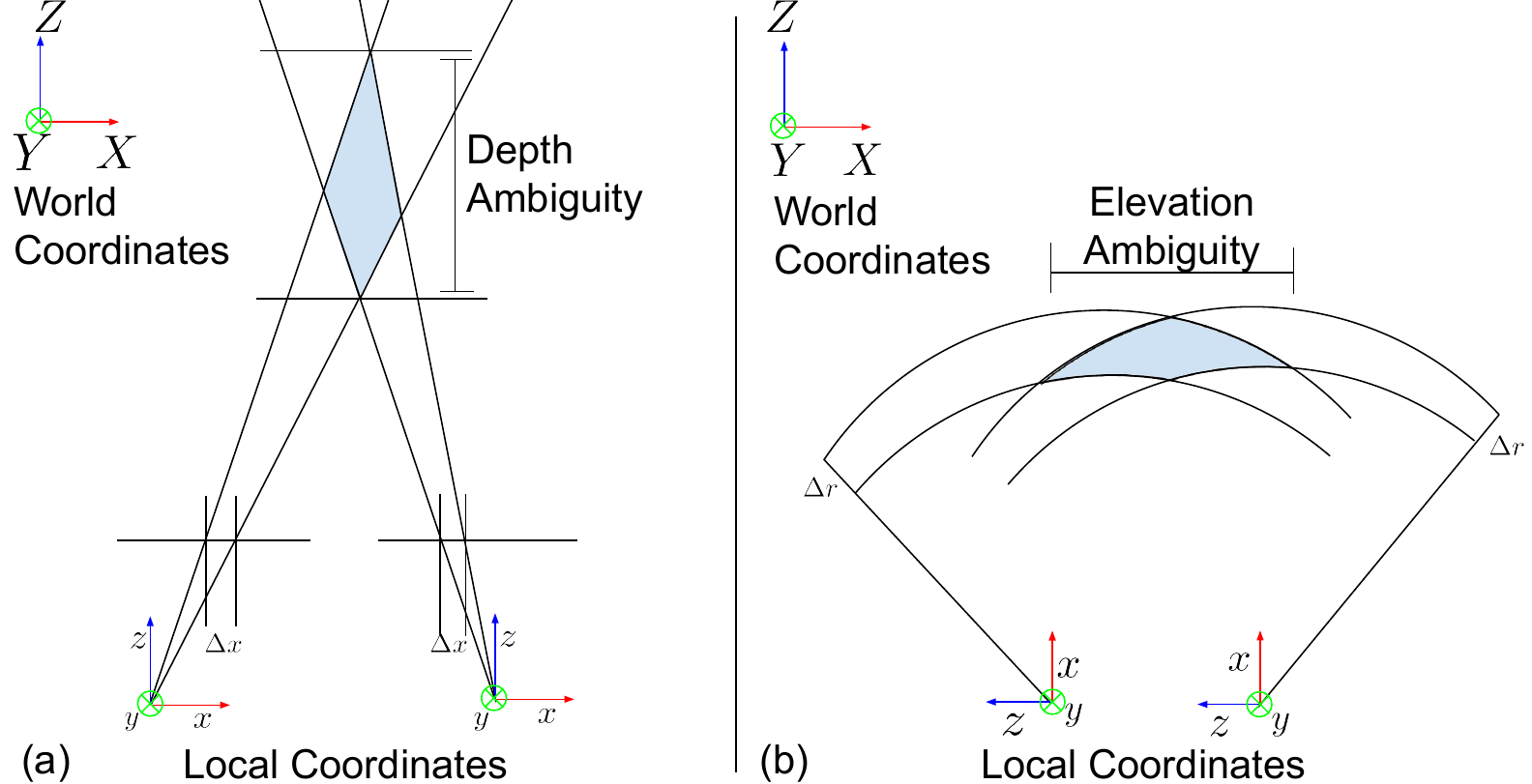}
   \caption{\textbf{Acoustic-Optical Measurement Ambiguities.}  (a) Two RGB measurements captured over a limited baseline struggle to localize a point along the depth-axis. (b) Two sonar measurements captured over a limited baseline struggle to localize a point along the x-axis. Because they have orthogonal ambiguities, RGB and sonar measurements are highly complementary.
} 
   \label{fig:Ambiguities}
   \vspace{-10pt}
\end{figure}

Our goal in this work is to reconstruct the 3D surface of an underwater object using a small collection of RGB and sonar measurements captured over a limited baseline. Specifically, we assume access to two datasets, $\mathcal{D}^{\text{cam}} = \{I_i^{\text{cam}}, P_i^{\text{son}} \}$ and $\mathcal{D}^{\text{son}} = \{I_i^{\text{son}}, P_i^{son} \}$, consisting of RGB/sonar images and their respective poses.

Given a large dataset captured over a sufficiently diverse range of poses (e.g., thousands of images captured from 360-degrees~\cite{qadri2023neural}), existing unimodal (camera-only/sonar-only) surface reconstruction methods are already effective~\cite{qadri2023neural,wang2021neus}. In this work, we focus on the small baseline operating conditions---pervasive in underwater robotics---where optical cameras record insufficient information to recover depth information (see~\cref{fig:Ambiguities}(a)) and imaging sonars record insufficient information to recover elevation information (see~\cref{fig:Ambiguities}(b)). 

Specifically, we introduce a physics-based multimodal inverse-differentiable-rendering framework that integrates information from both acoustic and optical sensors to generate accurate 3D reconstructions. 
Our approach automatically exploits the complementary information (elevation/range) provided by each sensor. 

Because our shared pool-based testing facility does not allow us to introduce turbidity, in this work we focus on the clear-water setting. Modeling the effects of light scattering in our forward model~\cite{sethuraman2023waternerf,levy2023seathru,ramazzina2023scatternerf,Chen2024dehazenerf} would likely improve our system's in-the-wild performance.

\section{Method}

\subsection{Acoustic-Optical NeuS}
\begin{figure}[t]
  \centering
  \Description{A diagram of the neural architecture that we use to represent the scene in our neural rendering framework.}
   \includegraphics[width=0.9\linewidth]{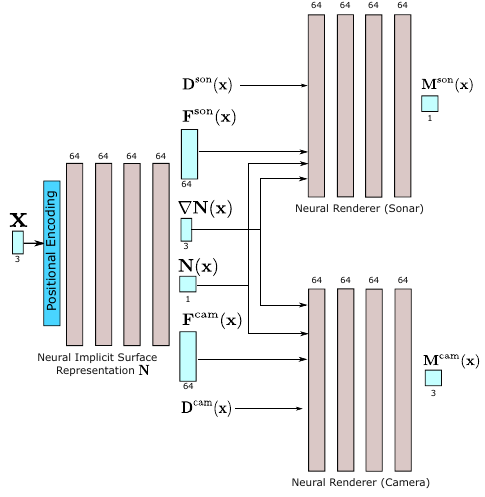}
   \caption{\textbf{AONeuS Reconstruction Framework.}  A shared surface geometry SDF network $\mathbf{N}$ is used in combination with rendering specific neural rendering modules. For each sampled point $\mathbf{x}$ along an acoustic or optical ray, $\mathbf{N}$ outputs its signed distance, its gradient as well as 2 features vectors $\mathbf{F}^{\text{son}}$ and $\mathbf{F}^{\text{cam}}$ all serving as input to their respective rendering networks. $\textbf{D}^{\text{son}}$ and $\textbf{D}^{\text{cam}}$ are respectively the directions of the acoustic and optical rays.}
   \label{fig:Architecture}
   \vspace{-10pt}
\end{figure}

Our AONeuS reconstruction framework is illustrated in~\cref{fig:Architecture}. 
Following~\citet{wang2021neus, qadri2023neural}, we represent the object's surface using a Signed Distance Function (SDF), $\mathbf{N}(\mathbf{x})$, which outputs the distance of each 3D point $\mathbf{x}=(\mathbf{X},\mathbf{Y},\mathbf{Z})$ to the nearest surface. Distinct from these works, we use two separate rendering neural networks ($\textbf{M}^{\text{cam}}$ and $\textbf{M}^{\text{son}}$) that approximate the optical and acoustic outgoing radiance at each spatial coordinate $\mathbf{x}$. This choice is motivated by the fact that different materials have different acoustic and optical reflectance properties. 
For example, glass is invisible to optical cameras but visible to imaging sonar, and PVC is invisible to imaging sonar but visible to optical cameras. 

In this work, we sample and sum points along acoustic and optical rays to approximate the rendering integrals defined by~\cref{renderinglossSonar} and~\cref{renderinglossCam}. Our rendering functions can be expressed as

\begin{align}
&    \hat{I}^{\text{son}}(r, \theta) = \sum_{\mathbf{x} \in \mathcal{A}_{p^{\text{son}}}}
\frac{1}{r(\mathbf{x})} T[\mathbf{x}] \alpha[\mathbf{x}] \mathbf{M}^{\text{son}}(\mathbf{x}),\text{ and }
\label{discreteImageFormationModelSonar}\\
&    \hat{I}^{\text{cam}}(x, y) = \sum_{\mathbf{x} \in \mathcal{R}_{p^{\text{cam}}}}  T[\mathbf{x}] \alpha[\mathbf{x}] \mathbf{M}^{\text{cam}}(\mathbf{x}),
\label{discreteImageFormationModelCamera}
\end{align}
\noindent where $\mathcal{A}_{p^{\text{son}}}$ is the set of sampled points along the acoustic arc at pixel $p^{\text{son}}$ and $\mathcal{R}_{p^{\text{cam}}}$ is the set of sampled points along optical ray passing through pixel $p^{\text{cam}}$. $ \mathbf{M}^{\text{son}}$ and $ \mathbf{M}^{\text{cam}}$ are the predicted radiances at $\mathbf{x}$.

The computation of the discrete transmittance and opacity terms in \cref{discreteImageFormationModelSonar} and \cref{discreteImageFormationModelCamera} requires sampling along both acoustic and optical rays. For any such spatial sample $\mathbf{x}_s$, (i.e., any point along an acoustic or optical ray), the discrete opacity at $\mathbf{x}_s$ can be approximated as
 \begin{align}
    \alpha[\mathbf{x}_s] = \max\left(\frac{\mathbf{\Phi}_q(\mathbf{N}(\mathbf{x}_{s}))-\mathbf{\Phi}_q(\mathbf{N}(\mathbf{x}_{s+1}))}{\mathbf{\Phi}_q(\mathbf{N}(\mathbf{x}_{s}))}, 0\right),
    \label{discreteopacity}
\end{align} 
where $\mathbf{\Phi}_q(x) = (1+e^{-qx})^{-1}$ is the Sigmoid function and $q$ is a trainable parameter. The discrete transmittance is modeled as 
\begin{align}
    T[\mathbf{x_s}] = \prod_{\mathbf{x_r} \;|\; r < s} (1-  \alpha[\mathbf{x_r}]).
\end{align}

\subsubsection{Loss Function}
Our loss function comprises the sonar and camera intensity losses: 
\begin{align}
& \mathcal{L}_{\text{int}}^{\text{son}} \equiv \frac{1}{\mathbf{N}_\mathcal{P^{\text{son}}}} \sum_{p \in \mathcal{P}^{\text{son}}}||\hat{I}(p) - I(p)||_1,\text{ and } \\
& \mathcal{L}_{\text{int}}^{\text{cam}} \equiv \frac{1}{\mathbf{N}_\mathcal{P^{\text{cam}}}} \sum_{p \in \mathcal{P}^{\text{cam}}}||\hat{I}(p) - I(p)||_1,
\end{align}
where $\mathcal{P}^{\text{cam}}$ and $\mathcal{P}^{\text{son}}$ is the set of sampled pixels in the camera and sonar images respectively. We additionally use the eikonal loss as an implicit regularization to encourage smooth reconstructions:
\begin{align}
    \mathcal{L}_{\text{eik}} \equiv \frac{1}{|\mathbf{X}|} \sum_{\mathbf{x}\in \mathbf{X} } (||\nabla \mathbf{N}(\mathbf{x})||_2 - 1)^2,
\end{align}
where $\mathbf{X}$ is the set of all sampled points. 

We also utilize an $\ell_1$ loss term as an additional prior term which biases the network towards reconstructions that minimize the total opacity of the scene (for example in cases where the object is on the seafloor and only specific sides can be imaged):
\begin{align}
    \mathcal{L}_{\text{reg}} \equiv \frac{1}{|\mathbf{X}|} \sum_{\mathbf{x} \in \mathbf{X}} || \alpha[\mathbf{x}]||_1.
\end{align}

Hence, our total loss is
\begin{align}
\mathcal{L}=\alpha(t)\mathcal{L}_{\text{int}}^{\text{son}}+(1-\alpha(t))\mathcal{L}_{\text{int}}^{\text{cam}}+\lambda_{\text{eik}}\mathcal{L}_{\text{eik}}+\lambda_{\text{reg}}\mathcal{L}_{\text{reg}}.
\label{loss_function}
\end{align}
The network is trained with the ADAM optimizer. 
\subsubsection{Weight Scheduling}\label{ssec:weighting}

The weights assigned to the sonar and camera intensity losses (respectively $\alpha(t)$ and $1-\alpha(t)$ in eq. \ref{loss_function}) impact the reconstruction quality as they determine which measurements the network should emphasize throughout training. 
We adopt a simple two-step weighting scheme:

\begin{align}
    \alpha(t) = \begin{cases}
    1 \text{ if } t < E_t, \\ 
    \lambda \text{ if } E_t < t < E_e .
    \end{cases}
    \label{weighting_eq}
\end{align}
In the early iterations, $t < E_t$, the sonar measurements are used exclusively 
and serve to "mask`` the object; i.e., update the weights of the SDF network $\mathbf{N}$ to bias it towards reconstructions in which the geometry of the object are better constrained in the depth direction. This process establishes an initialization for the later iterations.

In later iterations, $t >E_t $,  more emphasis is placed on the camera measurements. These measurements constrain the $x$ and $y$ directions and help resolve the elevation ambiguity inherent in sonar data. In this phase, sonar measurements receive less weight and act as a depth regularizer.
See section A.3 of the appendix for a comparison of our mixing procedure against different weighting schemes.

%% file: sec/4_experimental_results.tex
\section{Experimental Results}

In this section, we evaluate the AONeuS on both synthetic and experimentally captured data. Hyperparameters for our experiments can be found in section A.3 of the appendix material.

\subsection{Notation}
\begin{figure}[t!]
  \centering
   \Description{A figure that shows how we set up our simulations in Blender.}
   \includegraphics[width=0.85\linewidth]{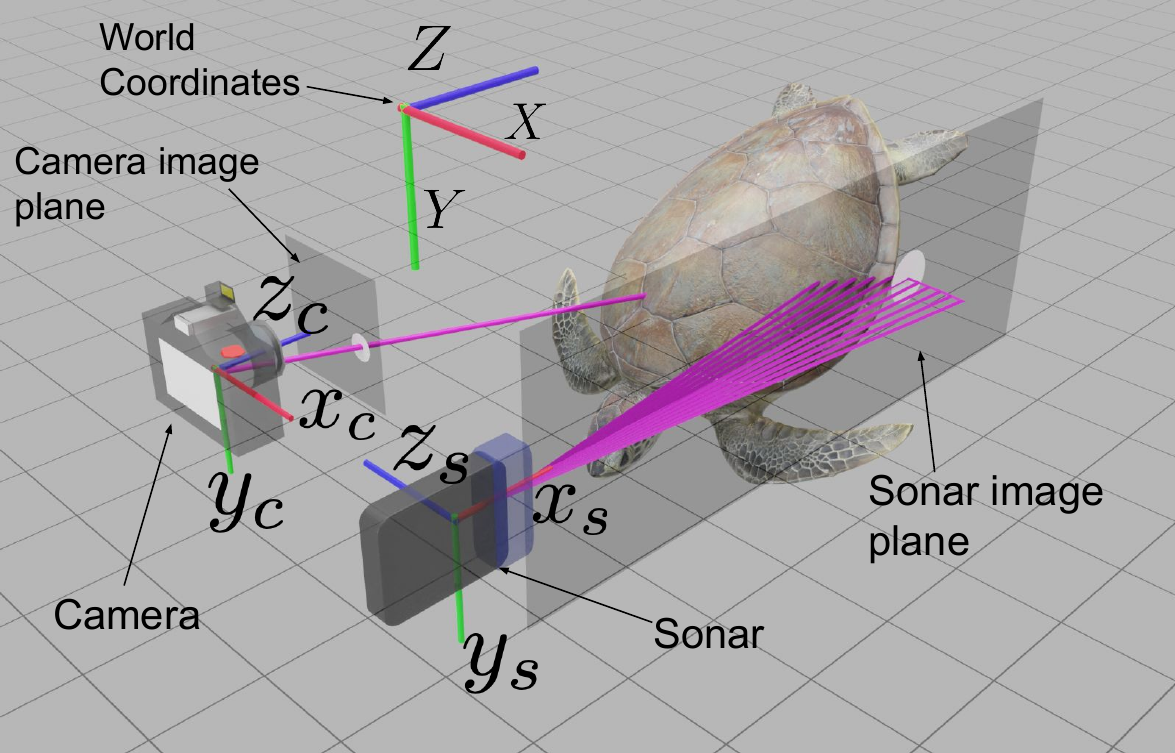}
   \caption{{\bf Simulation setup.} We visualize the orientations of the camera and the sonar relative to the scene, here a turtle, in our simulation. The optical axis of the camera, the $z$ axis of its own local coordinate frame, is aligned with the $Z$ axis of the world coordinate frame. The elevation axis of the sonar, the $z$ axis of its own local coordinate frame, is aligned with the $-X$ axis of the world coordinate frame. Additionally, we visualize the image planes of the camera and sonar, the 2D planes onto which they project the 3D scene points. }
   \label{fig:trajectory_vis}
   \vspace{-10pt}
\end{figure}

We use uppercase X,Y,Z to refer to the world coordinate system and lowercase x,y,z to refer to sensor-specific coordinate systems. See ~\cref{fig:trajectory_vis}.

\subsection{Results on Synthetic Data}
\label{ssec:simulation_experiments}

To generate synthetic measurements, we used a a custom-made sonar simulator as well as  Blender \cite{blender} for RGB measurements
to collect simulated sonar-camera datasets for various objects. 
The objects are assumed to be acoustically diffuse. We investigate the impact specular reflections have on the performance of our method in section A.1 of the appendix, which also lists the values of the simulation parameters we used.
The sonar and camera are approximately collocated, and are translated linearly over a short baseline along the $X$ axis of the world frame for a distance of \qty{1.2}{\metre} with the sonar's azimuthal plane parallel the YZ plane in the world frame. 
The sonar's azimuthal plane is oriented orthogonal to the direction of motion to ensure the trajectory was non-degenerate; multiple measurements captured from positions within the azimuthal plane of the sonar would be highly redundant and uninformative~\cite{negahdaripour2018application}. 

For each object, the trajectory is sub-sampled into smaller baselines: \qty{0.96}{\metre}, \qty{0.72}{\metre}, \qty{0.48}{\metre}, and \qty{0.24}{\metre} for analysis. 
We scaled the meshes so that the objects are approximately $\sim 1m$ in size and the sensors are placed about \qty{1.5}{\metre}--\qty{2}{\metre} away from the object. 
The elevation aperture of the sonar is $12^{\circ}$. 
We benchmark our method against two methods: NeuS~\cite{wang2021neus} and NeuSIS~\cite{qadri2023neural}, executing all methods 9 times with randomly initialized seeds. 
To ensure we had reasonable camera-only results, we provided NeuS with masks of the object. 
This information is not required by nor provided to NeuSIS and AONeuS. 

In~\cref{fig:SimManyObj}(a), we compare the reconstruction performance of all three techniques for a total of five scenes. 
We could observe that AONeus consistently reconstructs the scene geometry better than NeuS and NeuSIS. Further, we can also observe that NeuS (camera-only) incorrectly reconstructs the depth axis ($Z$-axis) whereas NeuSIS (sonar-only) can reconstruct only the depth-axis accurately. The proposed AONeus was able to recover underlying scene geometry along all the axes. 

\begin{table}[b!]
    \caption{\textbf{Metrics for synthetic turtle data.} Best metrics are bolded.}
    \label{table:turtle}
\vspace{-2mm}
    \begin{center}
    \resizebox{.9\linewidth}{!}{
\begin{tabular}{l|c|c|c|c}
\toprule
 &  & NeuS & NeuSIS & AONeuS \\
\hline
\multirow[c]{3}{*}{1.2m} & Chamfer $\boldsymbol{^\downarrow}$& $0.123 \pm 0.028$ & $0.130 \pm 0.013$ & $\mathbf{0.075} \pm 0.006$ \\
 & Precision$\boldsymbol{^\uparrow}$ & $0.653 \pm 0.095$ & $0.566 \pm 0.043$ & $\mathbf{0.862} \pm 0.042$ \\
 & Recall $\boldsymbol{^\uparrow}$& $0.526 \pm 0.134$ & $\mathbf{0.836} \pm 0.022$ & $0.825 \pm 0.056$ \\
 \hline
\multirow[c]{3}{*}{0.96m} & Chamfer $\boldsymbol{^\downarrow}$ & $0.139 \pm 0.024$ & $0.134 \pm 0.011$ & $\mathbf{0.079} \pm 0.005$ \\
 & Precision $\boldsymbol{^\uparrow}$& $0.602 \pm 0.076$ & $0.531 \pm 0.031$ & $\mathbf{0.840} \pm 0.017$ \\
 & Recall $\boldsymbol{^\uparrow}$& $0.470 \pm 0.132$ & $\mathbf{0.816} \pm 0.031$ & $0.807 \pm 0.017$ \\
 \hline
\multirow[c]{3}{*}{0.72m} & Chamfer $\boldsymbol{^\downarrow}$& $0.205 \pm 0.027$ & $0.135 \pm 0.011$ & $\mathbf{0.081} \pm 0.005$ \\
 & Precision $\boldsymbol{^\uparrow}$& $0.423 \pm 0.051$ & $0.537 \pm 0.062$ & $\mathbf{0.810} \pm 0.032$ \\
 & Recall $\boldsymbol{^\uparrow}$& $0.279 \pm 0.071$ & $0.768 \pm 0.029$ & $\mathbf{0.792} \pm 0.022$ \\
 \hline
\multirow[c]{3}{*}{0.48m} & Chamfer $\boldsymbol{^\downarrow}$& $0.249 \pm 0.045$ & $0.139 \pm 0.012$ & $\mathbf{0.088} \pm 0.006$ \\
 & Precision $\boldsymbol{^\uparrow}$& $0.337 \pm 0.062$ & $0.470 \pm 0.028$ & $\mathbf{0.791} \pm 0.023$ \\
 & Recall $\boldsymbol{^\uparrow}$& $0.189 \pm 0.074$ & $0.706 \pm 0.028$ & $\mathbf{0.770} \pm 0.023$ \\
 \hline
\multirow[c]{3}{*}{0.24m} & Chamfer $\boldsymbol{^\downarrow}$& $0.406 \pm 0.087$ & $0.146 \pm 0.009$ & $\mathbf{0.111} \pm 0.017$ \\
 & Precision $\boldsymbol{^\uparrow}$& $0.223 \pm 0.060$ & $0.450 \pm 0.028$ & $\mathbf{0.690} \pm 0.045$ \\
 & Recall $\boldsymbol{^\uparrow}$& $0.107 \pm 0.049$ & $0.587 \pm 0.042$ & $\mathbf{0.679} \pm 0.042$ \\
\bottomrule
\end{tabular}

    }
    \end{center}
    \vspace{-10pt}
\end{table}

In~\cref{fig:SimManyObj}(b), we show the results for the turtle mesh for various baselines. 
To visualize the ambiguities associated with camera and sonar modalities and the benefit of the fusion algorithm, we rendered the reconstructed meshes with a virtual camera pointing in $Y$-axis. Hence, the rendered images are projections of the reconstructed mesh on $ZX$-plane. 
As we decrease the baseline (top to bottom), for NeuS, we observe an increasing loss of features along depth direction: the back legs of the turtle are progressively lost and depth-reconstruction worsens with decreasing baselines.
For sonar-only methods, significant ambiguities along the elevation axis can be seen across all baselines: due to the limited translation of the sonar, the collected measurements are not enough to constrain and resolve the turtle shell adequately. 
Our framework AONeuS integrates orthogonal information from both imaging modalities to yield reconstructions of higher quality across all baselines: all features of the turtle including its shell and its back legs are clearly discernible. 
These observations are further supported by the quantitative analysis in \cref{table:turtle} where we report the mean and standard deviation of the Chamfer $L_1$ distance, precision, and recall of the reconstructions over nine trials. 
The results demonstrate that AONeuS outperforms the existing methods, particularly with reduced baselines. 
Note that recall of NeuSIS appears to be slightly better than AONeuS but that is only because the NeuSIS generates a large blob that covers most part of the object. 
The per-baseline quantitative results for the remaining four meshes can be found in section A.5 of the appendix. 

\begin{figure}[t]
  \centering
  \Description{A figure showing the experimental setup in real life that we used to capture data.}
   \includegraphics[width=0.8\linewidth]{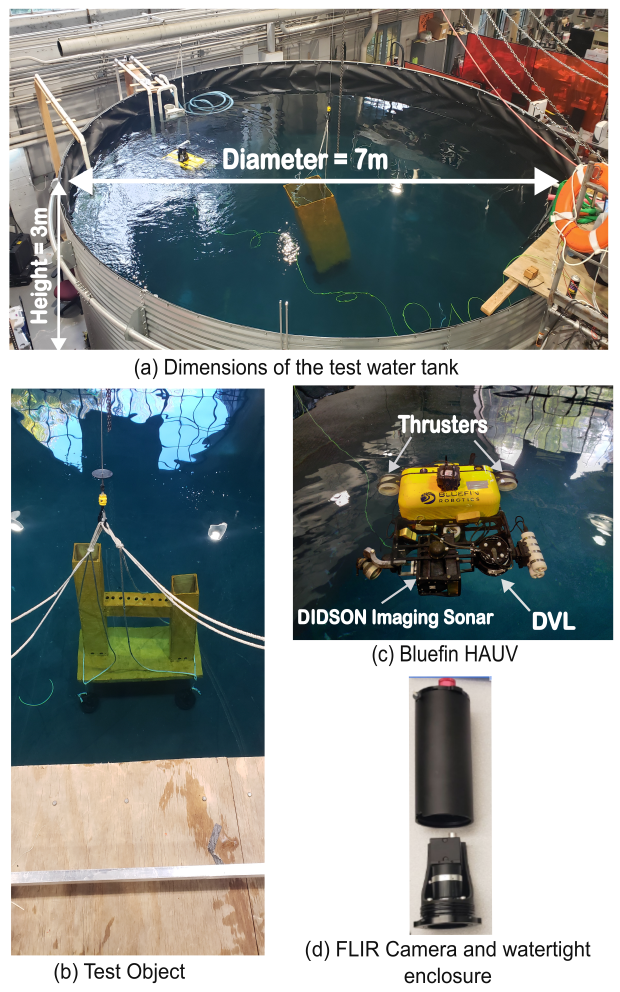}
   \caption{\textbf{Experimental hardware setup.} (a) Test water tank used to conduct the experiments and its dimensions. (b) Test object. (c) Bluefin Hovering Autonomous Underwater Vehicle (HAUV) and its mounted hardware (Didson imaging sonar and Doppler Velocity Log (DVL). (d) FLIR Blackfly S GigE camera used for image capture and its watertight enclosure.}
   \vspace{-10pt}
   \label{fig:exp_setup}
\end{figure}

\subsection{Results on Experimentally-Captured Data}\label{ssec:ExpResults}

We also perform real-world experiments on an object (\cref{fig:exp_setup}b) submerged in a water tank (\cref{fig:exp_setup}a). Please check the supplementary video for more visualizations of the setup. 
The object is made of standard wooden plywood of $\sim0.013$m thickness and has an acoustic impedence of $~2.5\times10^6$ $\text{kg}/\text{m}^2\text{s}$. Our object is also covered in a thin layer of insulating foam which increases its surface roughness and makes it more diffuse. 
We used a SoundMetrics DIDSON imaging sonar mounted on a Bluefin Hovering Autonomous Underwater Vehicle (HAUV) (\cref{fig:exp_setup}c) to capture two sonar datasets of the test object with two different elevation apertures $14^\circ$ and $28^\circ$. 
The sonar operates at a frequency of 1.8MHz corresponding to a wavelength of 0.833mm.\footnote{assuming the speed of sound in water to equal $\sim 1500 \text{m/s}$}
The vehicle uses an IMU and a Doppler Velocity Log (DVL) to measure sonar pose information. 
We asynchronously capture optical images of the same object using a FLIR Blackfly S GigE 5MP camera (\cref{fig:exp_setup}d) with camera pose information computed with COLMAP. 
To reduce the water-glass-air effect inside the camera, we used a checkerboard, underwater, to measure our camera’s underwater intrinsic parameters.
The sonar and camera trajectories were aligned post-capture. 
Similar to the simulation setup, both camera and sonar followed an approximately $1.2m$ {non-degenerate} linear trajectory 
($\sim120$ total sonar and RGB  measurements)
, which we later sub-sampled into the same $5$ baselines. 
Because any additional measurements captured along that trajectory are highly redundant, adding additional measurements (without added viewpoint diversity) has a limited effect on reconstruction accuracy.
We benchmarked our method against three algorithms: The COLMAP based sensor fusion method introduced in~\cite{kim3DReconstructionUnderwater2019}\footnote{COLMAP outputs a sparse pointcloud. Hence, a mesh was computed using the ball pivoting algorithm \cite{bernardini1999ball}.}, NeuS~\cite{wang2021neus}, and NeuSIS~\cite{qadri2023neural}. 
For each dataset and sensor baseline, we executed each method six times with randomly initialized seeds except that of ~\cite{kim3DReconstructionUnderwater2019}, which is deterministic. 

Qualitatively, we observe in \cref{fig:ExpResultH} that AONeuS outputs a more complete shape across baselines compared to sonar-only (NeuSIS) and camera-only (NeuS) only methods: the hole, two legs, and crossbar are clearly discernible. 
Conversely, when using only sonar, parts of the object are not well reconstructed as we can observe, for example, with the long leg with NeuSIS at $14^{\circ}$. 
Similarly, camera-only methods result in the loss of features such as the hole accompanied with significant introduced depth errors. 
We quantify the results in table 10 of the appendix, in section A.5, where we report the mean and standard deviation of the Chamfer $L_1$ distance, precision, and recall against the ground truth mesh computed over six trials with different random seeds for training. 
We observe that the fusion of the acoustic and optical signals generates higher quality reconstruction, even with very short baselines measuring only \qty{24}{\centi\metre}, as indicated by the mean value of each metric. 
When comparing AONeus with sonar-only methods (NeuSIS), we note that, despite the increased elevation ambiguity introduced by the $28^\circ$  elevation aperture, our technique is able to leverage camera information and its constraints in the $x$ and $y$ axes to resolve spatial locations that are otherwise under-constrained when solely relying on sonar. 
Techniques that rely on a camera only (NeuS) exhibit a decrease in performance as the sensor baseline is reduced. 
Complementing camera with sonar information introduces constraints in the depth direction easing the resolution of depth which is known to be difficult to resolve with limited camera motion. 
We additionally emphasize the standard deviation of the Chamfer distance over the random seeds: the fusion of both modalities result in outputs that are more robust to the randomness of the algorithm (i.e. network initialization, point samples, etc.).

%% file: sec/5_discussion.tex
\section{Discussion and Analysis}

\subsection{Distribution of per-Axis Errors}

In \cref{fig:per_axis_errors}, we visualize the per-axis deviations from the ground truth for the synthetic turtle scene at 0.24 m baseline. 
We compute per-axis deviations by first determining the closest vertex in the dense ground truth mesh and taking absolute differences in x, y, and z coordinates. 
We histogram these deviations along all three axes and show them along rows in the \cref{fig:per_axis_errors}. 
We have repeated this procedure for NeuS, NeuSIS, and AONeuS and show them along the columns.

From the data, we can observe (1) NeuS has large deviations along $Z$ axes, (2) NeuSIS has large deviations along $X$ axes. 
These results are consistent with the ambiguities associated with their respective measurement processes. 
AONeuS has low spread on all axes as it captures the best of both camera (NeuS) and sonar (NeuSIS) imaging modalities.
Further analysis of the reconstruction error can be found in section A.4 of the appendix.

\subsection{Multimodal Sensing is Better Conditioned}
The strong empirical performance of our multimodal reconstructions can be explained in terms of system conditioning.
Given point correspondences between measurements, it is far easier to triangulate a point using multimodal acoustic-optical measurements than camera-only or sonar-only measurements.

Previous works have analyzed the
conditioning of an imaging system consisting of an acoustic-optical stereo pair~\cite{negahdaripour2009opti}, e.g., a sonar (only) on the left and a camera (only) on the right. 
In this work, we follow a similar analysis to characterize a multimodal stereo pair, e.g., one co-located sonar and camera pair on the left and another co-located sonar and camera pair on the right. Our results follow closely from Negahdaripour's analysis of a stereo pair of 2D imaging sonars~\citet{negahdaripour2018application}; our stereo camera measurements introduce four additional linear constraints to the system's forward model.

Consider a point $P=[X,Y,Z]^T$ that is observed by an acoustic-optical sensor from two positions. 
The sonar's azimuthal plane is the $yz$ plane, in its own coordinate system. The camera's image plane is the $z=f$ plane, in its own coordinate system. Without loss of generality, assume the sensor's coordinate system at its initial location is the world coordinate system and its coordinate system at its second position is described by a rotation $\mathbf{R}$ and translation $\mathbf{t}=[t_x,t_y,t_z]$. That is, the coordinate of point $P$ in the new coordinate system is $P'=\mathbf{R}P+\mathbf{t}$. 

Under this model, the acoustic-optical sensor records 8 measurements:
\begin{align}
    &x_c=f\frac{[1,0,0]P}{[0,0,1]P}, &&y_c=f\frac{[0,1,0]P}{[0,0,1]P},\nonumber\\
    &R=\|P\|, &&\theta=\tan^{-1}{\Big (} \frac{[0,1,0]P}{[0,0,1]P}{\Big )}, \nonumber\\
    &x_c'=f\frac{\mathbf{r}_1^tP+t_x}{\mathbf{r}_3^tP+t_z}, &&y_c'=f\frac{\mathbf{r}_2^tP+t_y}{\mathbf{r}_3^tP+t_z},\nonumber\\
    &R'=\sqrt{\|R\|^2+\|t\|^2+2\mathbf{t}^t\mathbf{R}P}, &&\theta'=\tan^{-1}{\Big (} \frac{\mathbf{r}_2^tP+t_y}{\mathbf{r}_3^tP+t_z}{\Big )},
\end{align}
where $\mathbf{r_i}$ denotes the $i^{th}$ row of $\mathbf{R}$.

Following~\citet{negahdaripour2009opti,negahdaripour2018application}, we can turn each of these measurements into seven linear constraints and one non-linear constraint on $P$. 
\begin{align}
\mathbf{A}_{multi}P&=b\text{ and }\|P\|^2=R^2\text{ with }\nonumber\\ 
\mathbf{A}_{multi}&=\begin{bmatrix}
(-f,0,x_c)\\
(0,-f,y_c)\\
(0,-1,\tan(\theta))\\
x_c'\mathbf{r}_3^t-f\mathbf{r}_2^t\\
y_c'\mathbf{r}_3^t-f\mathbf{r}_2^t\\
\tan(\theta')\mathbf{r}_3^t-\mathbf{r}_2^t\\
\mathbf{t}^t\mathbf{R}
\end{bmatrix}
\text{ and }
b=\begin{bmatrix}
0\\
0\\
0\\
ft_x-x_c't_z\\
ft_y-y_c't_z\\
t_y-\tan(\theta')t_z\\
\frac{(R')^2-R^2-\|\mathbf{t}\|^2}{2}
\end{bmatrix}.
\end{align}

\begin{figure}[t]
  \centering
   \Description{A set of charts showing that the distribution of condition numbers for inverting a sonar-rgb imaging system is better (lower) than that of sonar or rgb alone.}
   \includegraphics[width=\linewidth]{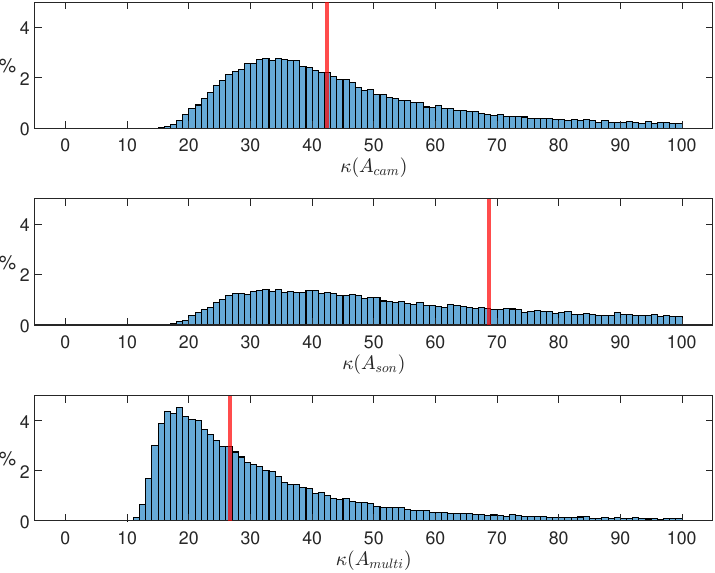}
   \caption{\textbf{System Conditioning.} Histograms of the condition numbers of the camera-only (top), sonar-only (middle), and multimodal (bottom) forward models. Median condition numbers are highlighted in red. The acoustic-optical multimodal forward model is generally better conditioned and easier to invert (triangulation).
   }
   \label{fig:CondtioningComparison}
   \vspace{-10pt}
\end{figure}

One can similarly form camera-only, $\mathbf{A}_{cam}$, and sonar-only, $\mathbf{A}_{son}$, forward models by considering only rows 1, 2, 4, and 5 and rows 3, 6, and 7, respectively, of $\mathbf{A}_{multi}$. By inverting these systems, one can triangulate $P$ in space. 
\balance 
Here we perform Monte Carlo sampling to compare the conditioning of $\mathbf{A}_{cam}$, $\mathbf{A}_{son}$, and $\mathbf{A}_{multi}$. We sample $P$ uniformly in a 1 $m^3$ cube centered at $(0,0,1.5)$ with edges parallel to the $x$, $y$, and $z$ axis; we assume $f=\qty{100}{\milli\metre}$; we sample $t_x$, $t_y$, and $t_z$ uniformly in the range $\qty{0}{\centi\metre}$ to $\qty{10}{\centi\metre}$; and we sample the yaw, pitch, and roll between measurements uniformly in the range \qty{-5}{\degree} to \qty{5}{\degree} .

For each realization of these parameters, we compute the condition number, $\kappa$, of $\mathbf{A}_{cam}$, $\mathbf{A}_{son}$, and $\mathbf{A}_{multi}$. We repeat this process $50,000$ times to form histograms, illustrated in \cref{fig:CondtioningComparison}. The condition number of the multimodal system is generally much lower and the system is thus easier to invert; multimodal triangulation is easier 
as each sensor does indeed contribute complementary constraints
.

\section{Limitations}

Currently, our approach is limited to clear-water settings and does not accurately model effects such as optical scattering and water absorption.  Hence, using more sophisticated optical physical models will be important when moving towards open-sea experiments. Similar to other differentiable rendering methods, our approach is too slow for online reconstruction (currently $\sim 30$ min) per scene on an Nvidia 3090 GPU. Enabling real-time reconstruction is an important direction for future work. Finally, we assume that our robot’s trajectory is nearly orthogonal to the sonar’s azimuthal plane: When the trajectory is within this plane additional sonar measurements provide little information. 

\section{Conclusion}
\label{sec:conclusion}
We have introduced and validated a multimodal inverse-differentiable-rendering framework for reconstructing 3D surface information from camera and sonar measurements.
Our framework combines camera and sonar information using a unified surface representation module and separate modality-specific appearance modules and rendering functions.
By extracting information from these complementary modalities, our framework is able to offer breakthrough underwater sensing capabilities for restricted baseline imaging scenarios.
We have demonstrated that AONeuS can accurately reconstruct the geometry of complex 3D objects from synthetic as well as noisy, real-world measurements captured over severely restricted baselines. 

While we demonstrate the first neural fusion of camera and sonar measurements, there are many interesting directions to explore this amalgamation. 
In~\cref{ssec:weighting}, we introduced a heuristic for weighing camera and sonar measurements. 
A structured way of combining the camera and sonar data, which is aware of the uncertainties~\cite{jiang2023fisherrf,goli2023} in the complementary imaging systems could result in faster convergence rates and better reconstructions. 

The sonar we have used in our implementations are forward-looking sonars. 
Fusion algorithms for side-scan sonars, synthetic-aperture sonars, sonars of different ranges and wavelengths, could be an interesting forward direction. 
Similarly, extending the technique for various geometries and materials including multi-object scenes, dynamic scenes, cluttered scenes and scattering media (murky water) would make AONeuS more practical. 
Finally, on-the-fly reconstructions could allow one to select the best next underwater view to improve reconstruction accuracy and further reduce the required baseline and acquisition time.

%% file: sec/7_ack.tex
\begin{acks}
The authors would like to thank Tianxiang Lin and Jui-Te Huang for their help with data collection; Sarah Friday for providing an animation showcasing the real experimental setup; and Shahriar Negahdaripour for helpful discussions, feedback, and guidance.  
M.Q.,~A.H.,~and M.K.~were supported in part by ONR grant N00014-21-1-2482. 
A.P.~was supported by NSF grant 2326904. 
K.Z.~and C.A.M.~were supported in part by AFOSR Young Investigator Program award no. FA9550-22-1-0208, ONR award no. N00014-23-1-2752, and a seed grant from SAAB, Inc. 
\end{acks}

%% file: sec/6_figure_only.tex
\begin{figure*}[!hb]
  \centering
   \Description{A set of qualitative results showing the better reconstruction quality of AONeuS compared to NeuS and NeuSIS in simulation.}
   \includegraphics[width=0.8\linewidth]{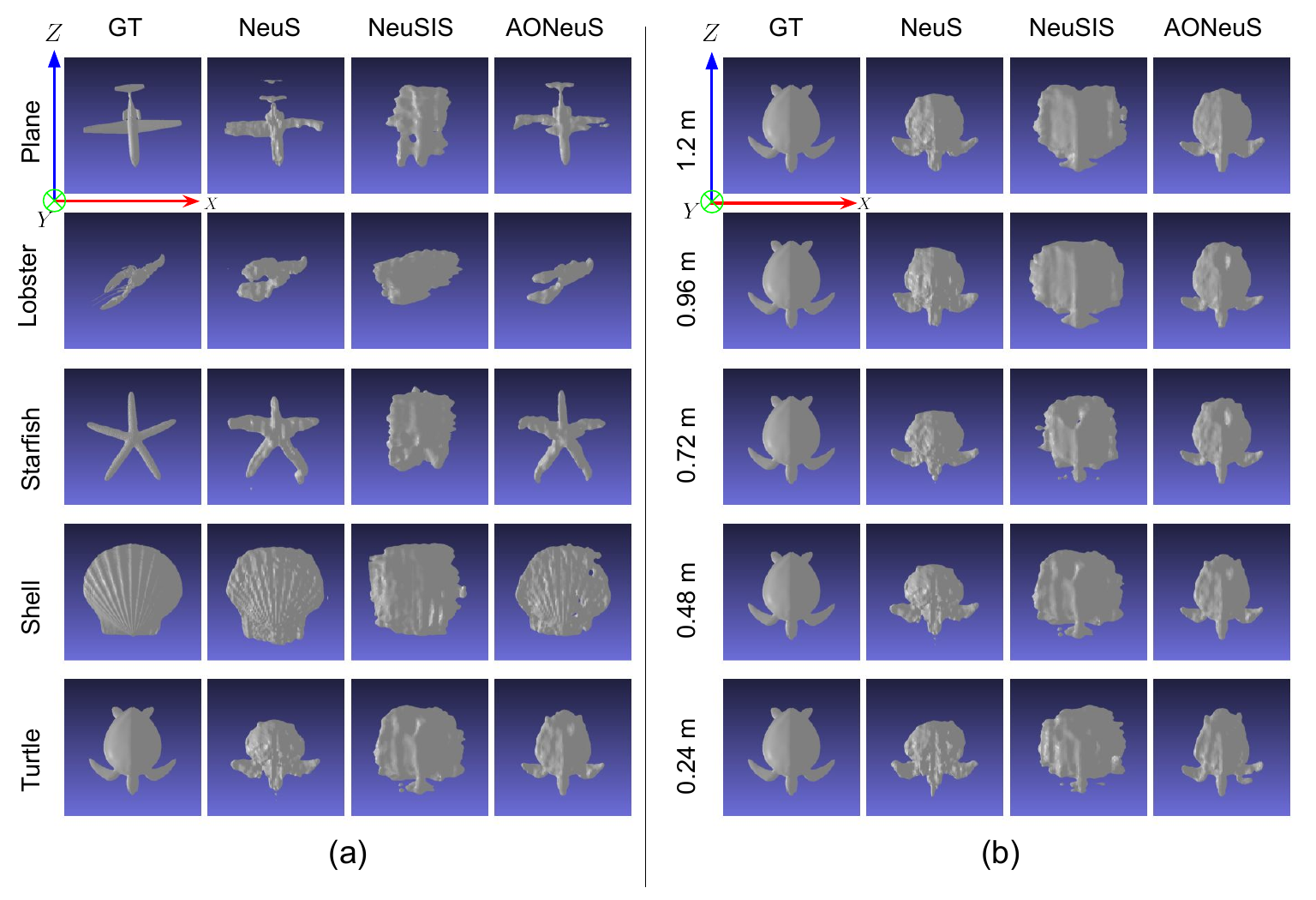}
   \caption{\textbf{Simulated Small-Baseline Results.} In (a), we visualize the reconstructions in simulation for all objects at 0.4x baseline, or 0.48 meters. In (b), we visualize the reconstructions in simulation for the turtle across all baseline settings.}
   \label{fig:SimManyObj}
\end{figure*}

\begin{figure*}[!ht]
  \centering
  \Description{A figure showing the better qualitative performance of AONeuS compared to NeuSIS, NeuS, and the previous state of the art on real experimental data.}
   \includegraphics[width=0.75\linewidth]{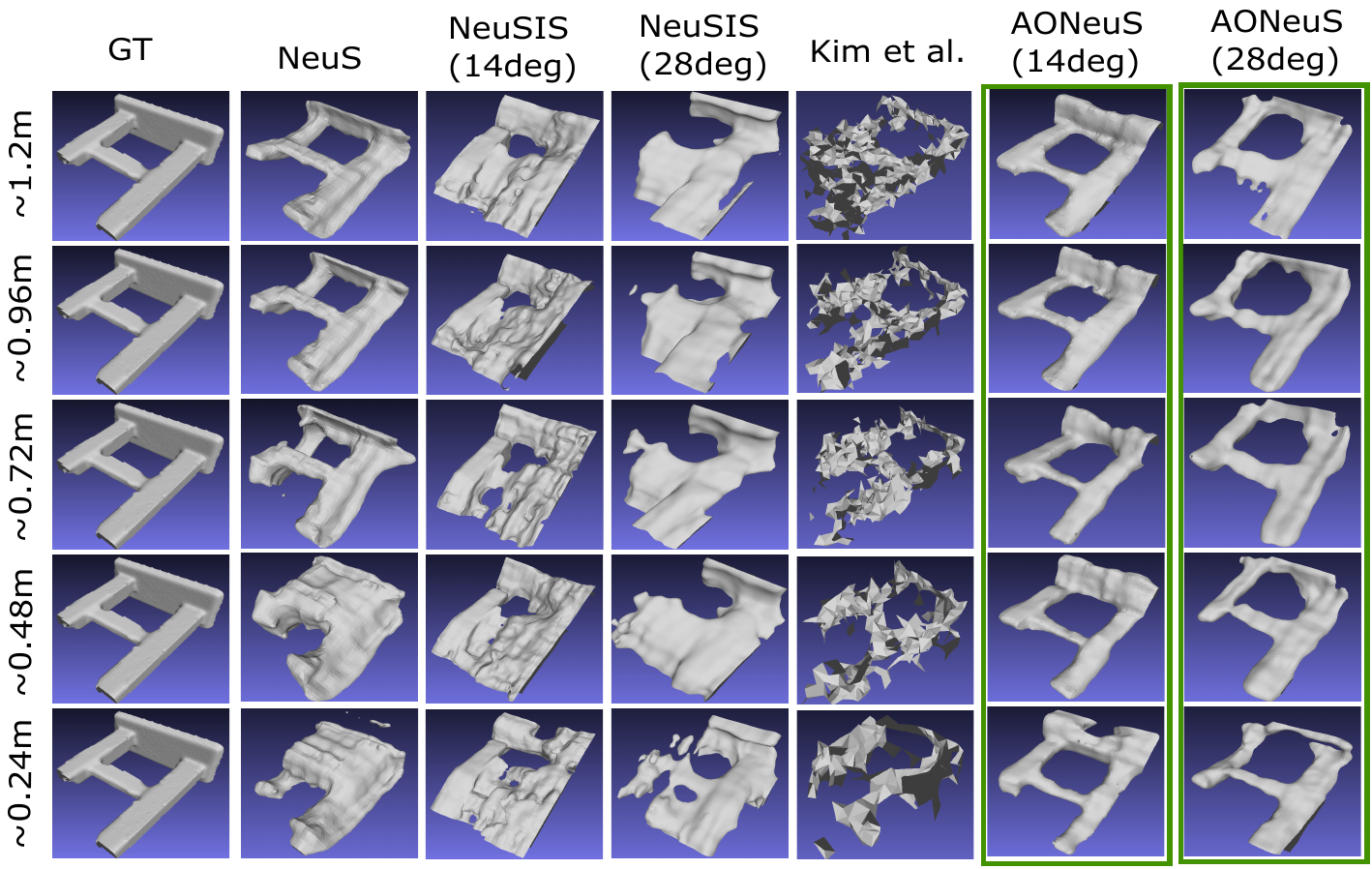}
   \caption{{\bf Experimental Results with ``h'' Object.} As the baseline diminishes, NeuS exhibits increasing amount of distortion along the depth direction as can seen at the intersection of the short piling and crossbar at the 0.72m and 0.96m baselines. NeuSIS similarly generates reconstructions with significant errors (for example, the long piling is poorly reconstructed with the $14^\circ$ elevation). Conversely, AONeuS consistently produces faithful reconstructions across a range of baselines. 
   }
   \label{fig:ExpResultH}
\end{figure*}
\begin{figure*}[!ht]
  \centering
   \Description{A set of charts showing that AONeuS has a better depth error distribution than NeuS and a better horizontal error distribution than NeuSIS.}
   \includegraphics[width=0.45\linewidth]{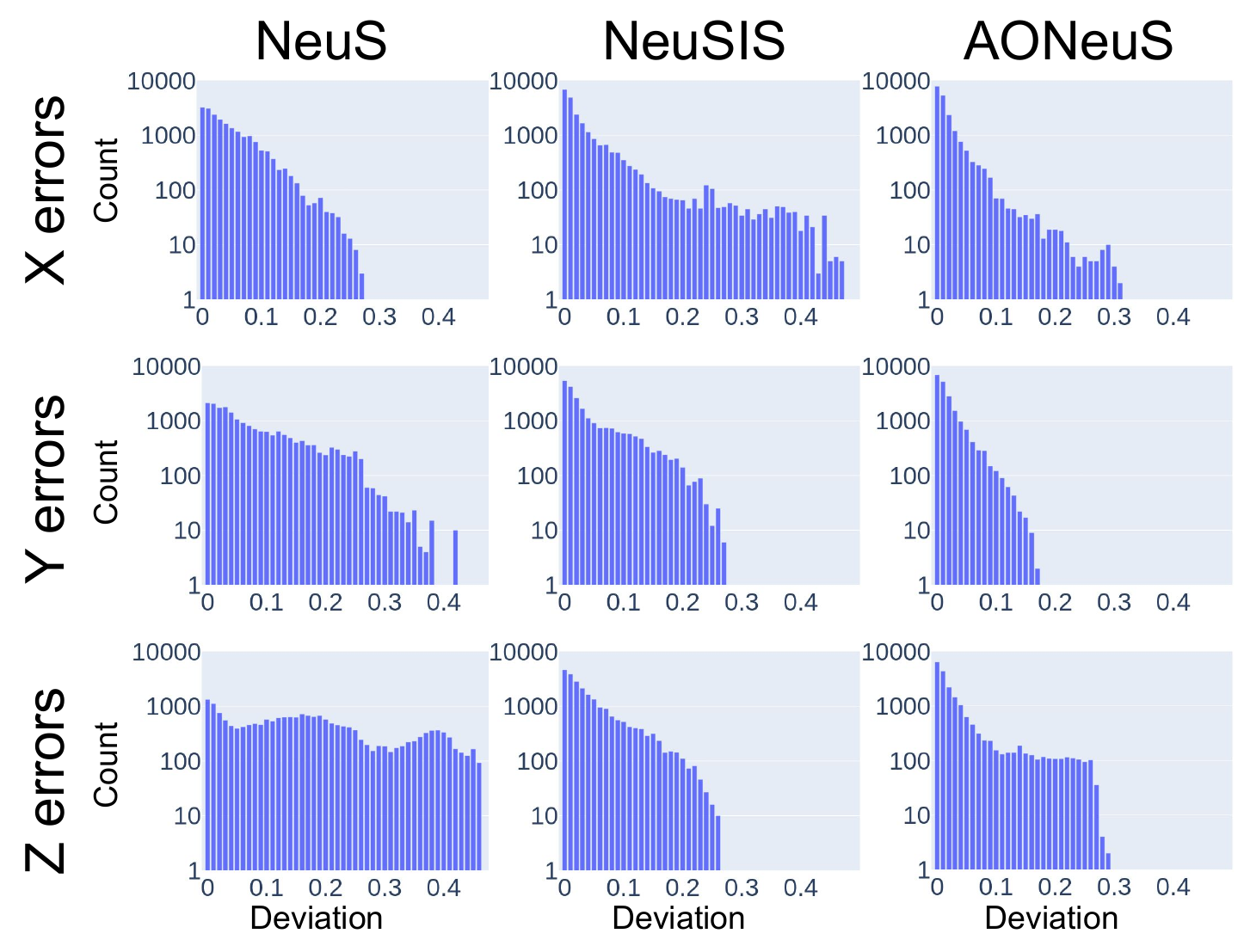}
   \caption{\textbf{Per-axis error distributions.} At 0.2x baseline for the turtle example, we plot the distributions of deviations from the ground truth mesh along all three axes for NeuS, NeuSIS, and AONeuS reconstructions. 
   The NeuS reconstruction has larger $Z$ errors, noticeable from the long tail, where as NeuSIS reconstruction has larger $X$ errors. 
   AONeuS has tighter distributions along all three axes compared to NeuS or NeuSIS showing that the proposed technique takes the best of both of the imaging modalities.}
   \label{fig:per_axis_errors}
\end{figure*}

%% file: sec/X_suppl.tex
\subsection{Effect of Acoustic Specularity on the Reconstruction Accuracy}
\label{simspecularanalysis}
\begin{figure}[h!]
    
  \Description{A graphic figure showing the sonar imaging formation model for different incidence angles.}
  \includegraphics[width=0.45\linewidth]{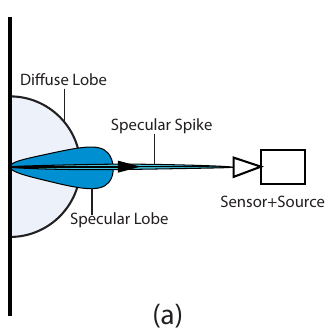}
  \includegraphics[width=0.45\linewidth]{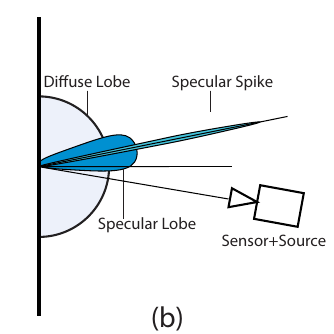}%
  \caption{\textbf{Comparing the received intensity for different incidence angles.} In (a), the sensor receives all components of the reflection: the diffuse lobe, specular lobe, and the specular spike. In (b), the sensor only receives the specular lobe and the diffuse lobes, because those reflection components have portions which are reflected under the normal to the surface, where the sensor is, whereas the specular spike is completely reflected above the normal of the surface, away from the sensor.}
  \label{fig:reflection_geometry}

\end{figure}

We investigate the reconstruction accuracy of AONeuS when imaging acoustically-specular objects. Consider the following sonar measurement formation model~\cite{langerBuildingQualitativeElevation1991} that has a diffuse and specular component, 
\begin{equation}
    I_r = \underbrace{C_{dl} \cos \alpha}_{\text{diffuse}} + \underbrace{C_{sl} G(\alpha) \frac{1}{\cos \alpha} \exp \left(-\frac{\alpha^2}{2\sigma_{\alpha}^2}\right)}_{\text{specular}},
    \label{langersonarimageformationmodel}
\end{equation}
where $I_r$ is the intensity of the reflection, $\alpha$ is the angle of incidence, $C_{dl}$ and $C_{sl}$ represent how strong the diffuse and specular components are, relatively, $\sigma_{\alpha}$ is the standard deviation of slope of the microfacet distribution which models the surface roughness, and $G(\alpha)$ is a geometric attenuation factor~\cite{nayarSurfaceReflectionPhysical1991} which represents how the microfacets might occlude each other. For sonar, 
\begin{equation}
    G(\alpha) = \min (1, 2\cos^2(\alpha)).
\end{equation}
We note that this is a modification of the extension of the Torrance-Sparrow reflection model~\cite{torranceTheoryOffSpecularReflection1967} developed in~\cite{nayarSurfaceReflectionPhysical1991} that excludes the direct specular spike component of the reflection and retains only the diffuse and specular lobe components of the reflection. The rationale is that for a sonar, the wave source and receiver are at the same position, so the sensor receives only reflected in a direction towards it. Therefore, the receiver will only receive the specular spike if the surface is close to perfectly normal to the sensor, which rarely happens in practice. The reflection geometry is described in more detail in ~\cref{fig:reflection_geometry}.
For the experiments in section 6.2, we set $C_{dl}=1$ and $C_{sl}=0$.

\begin{figure}[h!]
  \centering
  \Description{A figure showing the qualitative performance of AONeuS as the specularity of the object is changed, in simulation.}
   \includegraphics[width=0.8\linewidth]{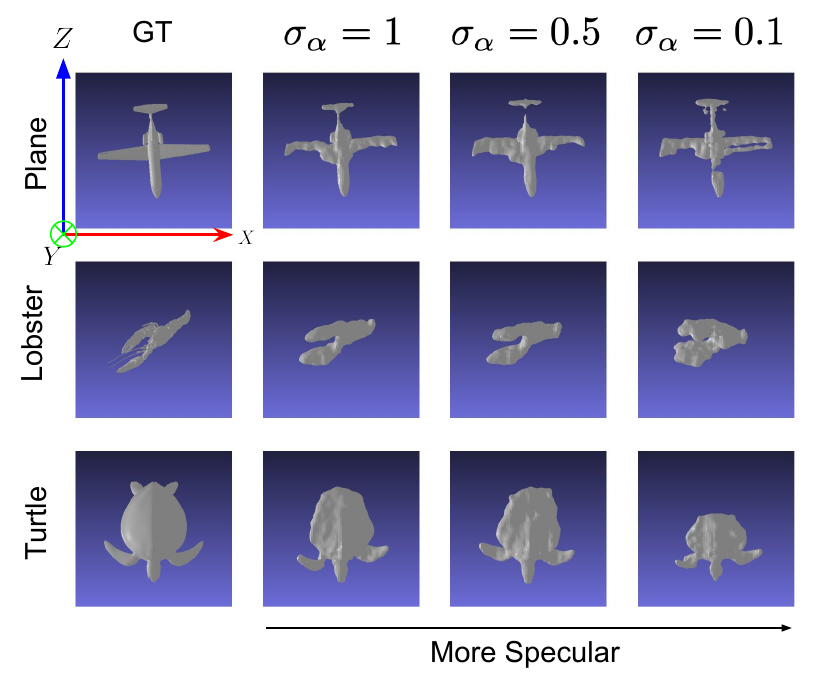}
   \caption{{\bf Reconstructions by AONeuS from RGB images and specular imaging sonar measurements.} As the width of the specular lobe decreases, i.e. as $\sigma_{\alpha}$ decreases, we observe that the performance of AONeuS decreases. 
   } \label{fig:specular_qualitative}
\end{figure}

In the following experiments, we focus on the specular component of eq.\ref{langersonarimageformationmodel} (i.e. set $C_{dl}=0,C_{sl}=1$) and investigate the effect of varying the parameter $\sigma_\alpha$ (which models the width of the specular lobe for purely acoustic specular reflections) on the reconstruction quality of AONeuS. All experiments were ran on the 0.24 m baseline.  
We visualize the qualitative results of the experiments in~\cref{fig:specular_qualitative}, which show that AONeuS can generate accurate reconstructions even from specular imaging sonar data. Of note is that as the width of the specular lobe decreases and becomes more narrow, the quality of the reconstruction decreases, implying that performing 3D reconstruction becomes more difficult. 
The qualitative trends are also validated by the quantitative results in~\cref{table:airplane_specular} to ~\cref{table:turtle_specular}. Here, we note that for all three meshes we consider, AONeuS performs best when $\sigma_{\alpha} = 1$, or when the specular lobe is the widest, and for certain geometries, like the turtle, the reconstruction performance degrades  at $\sigma_{\alpha}=0.1$.


\begin{table}[h!]
\caption{Quantitative metrics, airplane mesh, specular data.}
\label{table:airplane_specular}
\resizebox{.9\linewidth}{!}{
\begin{tabular}{l|c|c|c}
\toprule
 & Chamfer L1 $\boldsymbol{\downarrow}$ & Precision $\boldsymbol{\uparrow}$ & Recall $\boldsymbol{\uparrow}$ \\
\midrule
$\sigma_{\alpha} = 1.0$ & $0.141 \pm 0.026$ & $0.558 \pm 0.059$ & $0.667 \pm 0.058$ \\
$\sigma_{\alpha} = 0.5$ & $0.150 \pm 0.020$ & $0.503 \pm 0.064$ & $0.649 \pm 0.039$ \\
$\sigma_{\alpha} = 0.1$ & $0.149 \pm 0.027$ & $0.445 \pm 0.079$ & $0.629 \pm 0.054$ \\
\bottomrule
\end{tabular}
}

\end{table}

\begin{table}[h!]
\caption{Quantitative metrics, lobster mesh, specular data.}
\label{table:lobster_specular}
\resizebox{.9\linewidth}{!}{
\begin{tabular}{l|c|c|c}
\toprule
 &  Chamfer L1  $\boldsymbol{\downarrow}$ & Precision $\boldsymbol{\uparrow}$ & Recall $\boldsymbol{\uparrow}$ \\
\midrule
$\sigma_{\alpha} = 1.0$ & $0.186 \pm 0.053$ & $0.440 \pm 0.142$ & $0.439 \pm 0.174$ \\
$\sigma_{\alpha} = 0.5$ & $0.183 \pm 0.035$ & $0.468 \pm 0.113$ & $0.441 \pm 0.120$ \\
$\sigma_{\alpha} = 0.1$ & $0.206 \pm 0.040$ & $0.329 \pm 0.077$ & $0.473 \pm 0.142$ \\
\bottomrule
\end{tabular}
}

\end{table}

\begin{table}[h!]
\caption{Quantitative metrics, turtle mesh, specular data.}
\label{table:turtle_specular}
\resizebox{.9\linewidth}{!}{
\begin{tabular}{l|c|c|c}
\toprule
 &  Chamfer L1  $\boldsymbol{\downarrow}$ & Precision $\boldsymbol{\uparrow}$ & Recall $\boldsymbol{\uparrow}$ \\
\midrule
$\sigma_{\alpha} = 1.0$ & $0.116 \pm 0.018$ & $0.686 \pm 0.045$ & $0.691 \pm 0.045$ \\
$\sigma_{\alpha} = 0.5$ & $0.104 \pm 0.009$ & $0.717 \pm 0.048$ & $0.720 \pm 0.045$ \\
$\sigma_{\alpha} = 0.1$ & $0.191 \pm 0.027$ & $0.533 \pm 0.079$ & $0.615 \pm 0.061$ \\
\bottomrule
\end{tabular}
}
\end{table}

\newpage
\subsection{Hyperparameters}
\begin{table}[h!]
\caption{
List of hyperparameters.
}
\label{table:hyperparameters}
    \begin{center}
    \resizebox{1\linewidth}{!}{
        \begin{tabular}{c|c|c|c}
        \toprule 
            Parameter &\makecell{Sonar dataset 1 \\ $14^\circ$ elevation angle} & \makecell{Sonar dataset 2 \\ $28^\circ$ elevation angle} & Simulation\\
            \hline
            $E_t$ & 4000 & 4000 & 2000 \\ 
            $E_e$ & 8000 & 8000 & 5000\\ 
            $\lambda$ & 0.3 & 0.3& 0.3\\ 
            $\lambda_{\text{eik}}$ & $0.01$ & $0.1$ & 0.1\\ 
            $\lambda_{
\text{reg}}$ & $0.1$ & $1$ &  0\\ 
            \toprule
        \end{tabular}
    }
    \end{center}

\end{table}


\subsection{Ablation Study (Weighting Schemes)}
\begin{table}[h!]
    \caption{Various weighting scheme. Experimenting on the real object with 0.24m baseline. Values are averaged over 6 trials. \textbf{Constant}: fixed weights. \textbf{Linear}: Weights change linearly from initial to end values. \textbf{Step}: Weights are switched from start to end value at iteration $E_t$.}
    \label{table:ablationReal}
    \begin{center}
    \resizebox{1\linewidth}{!}{
        \begin{tabular}{c|c|c||c|c|c}
        \toprule
        \multicolumn{6}{c}{Camera +  sonar at $14^\circ$ elevation} \\
        \hline 
        Mode & $\alpha(t)$ start & $\alpha(t)$ end & Chamfer L1 $\boldsymbol{^\downarrow}$ & Precision $\boldsymbol{^\uparrow}$ & Recall $\boldsymbol{^\uparrow}$\\
        \hline 
        Constant & 0.5 & 0.5 & $0.120\pm 0.026$& $0.586\pm 0.055$& $0.624\pm0.124$\\ 
        Constant & 0.7 & 0.7 & $0.141\pm0.046$& $0.582\pm0.078$&$0.524\pm0.133$ \\ 
        Constant & 0.3 & 0.3 &  $0.093 \pm 0.009$&  $0.626\pm  0.049$& $0.741\pm0.035$\\ 
        Linear &  1  & 0  & $0.181\pm0.036$& $0.513\pm0.040$& $0.388\pm0.088$ \\ 
        Linear &  0  & 1  & $0.108\pm0.019$ & $0.597\pm0.056$& $0.670\pm0.092$\\ 
        Step & 0 & 0.7 & $0.113\pm0.01$0&$0.542\pm0.042$ &  $0.695\pm0.014$\\
        \textbf{Step (Ours)} & 1 & 0.3 & $\textbf{0.085}\pm0.009$ & $\textbf{0.706}\pm 0.063$& $\textbf{0.758}\pm0.041$\\
        \hline
         \multicolumn{6}{c}{Camera + sonar at $28^\circ$ elevation}   \\
         \hline
        Mode & $\alpha(t)$ start & $\alpha(t)$ end & Chamfer L1 $\boldsymbol{^\downarrow}$ & Precision $\boldsymbol{^\uparrow}$ & Recall $\boldsymbol{^\uparrow}$\\
        \hline 
        Constant & 0.5 & 0.5 & $0.131\pm0.014$& $0.437\pm0.048$& $0.604\pm0.061$ \\ 
        Constant & 0.7 & 0.7 & $0.121\pm0.008$& $0.513\pm0.038$& $0.585\pm0.036$\\ 
        Constant & 0.3 & 0.3 & $0.141\pm0.010$& $0.380\pm0.042$&$0.595\pm0.046$\\ 
        Linear &  1  & 0  & $0.140\pm0.027$ & $0.498\pm0.076$& $0.590\pm0.101$\\ 
        Linear &  0  & 1  & $0.144\pm0.015$  & $0.388\pm0.068$&$0.594\pm0.098$ \\ 
        Step &0 & 0.7 & $0.139\pm0.015$ & $0.454\pm0.068$& $0.572\pm0.086$\\
        \textbf{Step (Ours)} & 1  & 0.3 & $\textbf{0.108}\pm0.005$ & $\textbf{0.538}\pm0.022$& $\textbf{0.695}\pm 0.036$\\
        \hline
        \end{tabular}
    }
    \end{center}

\end{table}

\subsection{Variance of the Density Field Over Realizations of the Algorithm}
\begin{figure}[H]
  \Description{A figure showing the the more constrained of the distribution of the reconstructions of AONeuS versus those of NeuS and NeuSIS.}
  \centering
   \includegraphics[width=0.9\linewidth]{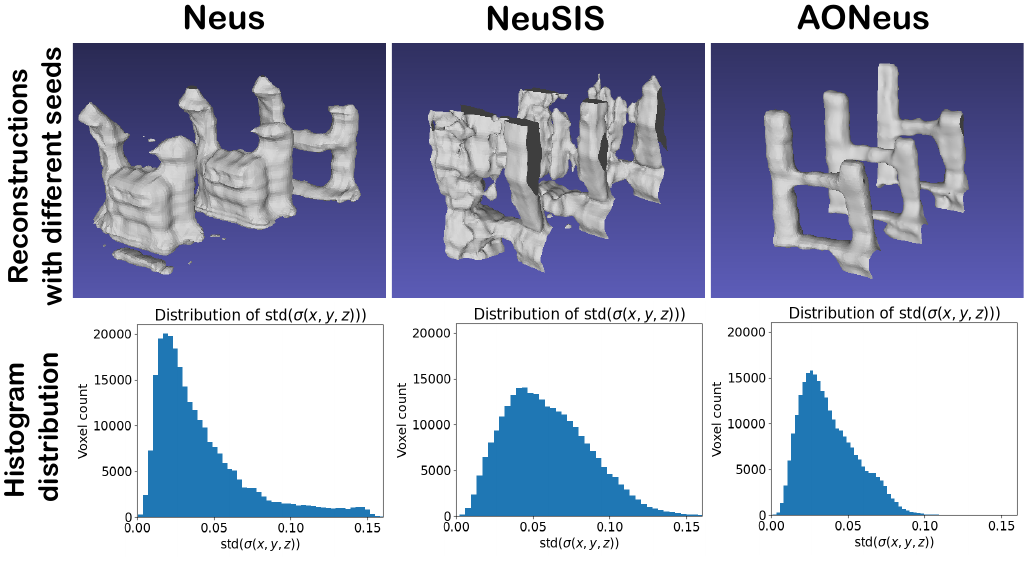}
   \caption{{\bf Distribution of Reconstructions of the Real Object.} Top: Different Reconstructions from different random seeds for the 0.24m baseline with the sonar elevation at $14^\circ$.  Bottom: Histogram distribution of the standard deviation of the density field vs. voxel count. Neus's distribution is heavily tailed while NeuSIS's distribution exhibits a large mean and variance. AONeuS's distribution is light-tailed with a small mean and therefore it is better constrained. 
   } \label{fig:VarianceVisualizationReal}
\end{figure}


\newpage
\subsection{Additional tables}
Tables~\ref{table:airplane} to~\ref{table:shell} provide additional quantitative metrics for our synthetic experiments.  

\begin{table}[hb!]
\caption{Quantitative metrics for the airplane mesh.}
\label{table:airplane}
\resizebox{.9\linewidth}{!}{
\begin{tabular}{l|c|c|c|c}
\toprule
 &  & NeuS & NeuSIS & AONeuS \\
\midrule
\multirow[c]{3}{*}{1.2m} & Chamfer $\boldsymbol{\downarrow}$ & $\mathbf{0.112} \pm 0.018$ & $0.197 \pm 0.011$ & $0.117 \pm 0.014$ \\
 & Precision $\boldsymbol{\uparrow}$ & $\mathbf{0.652} \pm 0.058$ & $0.295 \pm 0.019$ & $0.582 \pm 0.053$ \\
 & Recall $\boldsymbol{\uparrow}$ & $0.650 \pm 0.044$ & $0.643 \pm 0.025$ & $\mathbf{0.741} \pm 0.028$ \\
 \hline
\multirow[c]{3}{*}{0.96m} & Chamfer $\boldsymbol{\downarrow}$ & $0.144 \pm 0.021$ & $0.200 \pm 0.019$ & $\mathbf{0.134} \pm 0.016$ \\
 & Precision $\boldsymbol{\uparrow}$ & $0.559 \pm 0.045$ & $0.291 \pm 0.014$ & $\mathbf{0.575} \pm 0.017$ \\
 & Recall $\boldsymbol{\uparrow}$ & $0.579 \pm 0.042$ & $0.650 \pm 0.043$ & $\mathbf{0.697} \pm 0.027$ \\
 \hline
\multirow[c]{3}{*}{0.72m} & Chamfer $\boldsymbol{\downarrow}$ & $0.146 \pm 0.021$ & $0.200 \pm 0.016$ & $\mathbf{0.141} \pm 0.023$ \\
 & Precision $\boldsymbol{\uparrow}$ & $0.554 \pm 0.052$ & $0.289 \pm 0.029$ & $\mathbf{0.558} \pm 0.035$ \\
 & Recall $\boldsymbol{\uparrow}$ & $0.599 \pm 0.039$ & $0.629 \pm 0.067$ & $\mathbf{0.689} \pm 0.048$ \\
 \hline
\multirow[c]{3}{*}{0.48m} & Chamfer $\boldsymbol{\downarrow}$ & $0.174 \pm 0.016$ & $0.199 \pm 0.012$ & $\mathbf{0.146} \pm 0.033$ \\
 & Precision $\boldsymbol{\uparrow}$ & $0.468 \pm 0.039$ & $0.287 \pm 0.047$ & $\mathbf{0.533} \pm 0.087$ \\
 & Recall $\boldsymbol{\uparrow}$ & $0.516 \pm 0.040$ & $0.569 \pm 0.076$ & $\mathbf{0.668} \pm 0.044$ \\
 \hline
\multirow[c]{3}{*}{0.24m} & Chamfer $\boldsymbol{\downarrow}$ & $0.223 \pm 0.046$ & $0.182 \pm 0.011$ & $\mathbf{0.166} \pm 0.034$ \\
 & Precision $\boldsymbol{\uparrow}$ & $0.341 \pm 0.090$ & $0.358 \pm 0.042$ & $\mathbf{0.451} \pm 0.103$ \\
 & Recall $\boldsymbol{\uparrow}$ & $0.413 \pm 0.072$ & $0.555 \pm 0.069$ & $\mathbf{0.644} \pm 0.045$ \\
\bottomrule
\end{tabular}
}

\end{table}

\begin{table}[hb!]
\caption{Quantitative metrics for the lobster mesh.}
\label{table:lobster}
\resizebox{.9\linewidth}{!}{
\begin{tabular}{l|c|c|c|c}
\toprule
 &  & NeuS & NeuSIS & AONeuS \\
\midrule
\multirow[c]{3}{*}{1.2m} & Chamfer $\boldsymbol{\downarrow}$ & $0.147 \pm 0.017$ & $0.187 \pm 0.012$ & $\mathbf{0.105} \pm 0.019$ \\
 & Precision $\boldsymbol{\uparrow}$ & $0.448 \pm 0.073$ & $0.328 \pm 0.020$ & $\mathbf{0.592} \pm 0.110$ \\
 & Recall $\boldsymbol{\uparrow}$ & $0.454 \pm 0.100$ & $0.460 \pm 0.037$ & $\mathbf{0.626} \pm 0.090$ \\
 \hline
\multirow[c]{3}{*}{0.96m} & Chamfer $\boldsymbol{\downarrow}$ & $0.167 \pm 0.027$ & $0.203 \pm 0.014$ & $\mathbf{0.142} \pm 0.074$ \\
 & Precision $\boldsymbol{\uparrow}$ & $0.394 \pm 0.070$ & $0.302 \pm 0.025$ & $\mathbf{0.534} \pm 0.230$ \\
 & Recall $\boldsymbol{\uparrow}$ & $\mathbf{0.398} \pm 0.106$ & $0.445 \pm 0.033$ & $0.533 \pm 0.228$ \\
 \hline
\multirow[c]{3}{*}{0.72m} & Chamfer $\boldsymbol{\downarrow}$ & $0.191 \pm 0.019$ & $0.225 \pm 0.030$ & $\mathbf{0.128} \pm 0.021$ \\
 & Precision $\boldsymbol{\uparrow}$ & $0.357 \pm 0.053$ & $0.275 \pm 0.040$ & $\mathbf{0.533} \pm 0.112$ \\
 & Recall $\boldsymbol{\uparrow}$ & $0.417 \pm 0.077$ & $0.390 \pm 0.040$ & $\mathbf{0.576} \pm 0.105$ \\
 \hline
\multirow[c]{3}{*}{0.48m} & Chamfer $\boldsymbol{\downarrow}$ & $0.237 \pm 0.039$ & $0.243 \pm 0.019$ & $\mathbf{0.143} \pm 0.015$ \\
 & Precision $\boldsymbol{\uparrow}$ & $0.321 \pm 0.062$ & $0.257 \pm 0.012$ & $\mathbf{0.523} \pm 0.069$ \\
 & Recall $\boldsymbol{\uparrow}$ & $0.392 \pm 0.091$ & $0.370 \pm 0.017$ & $\mathbf{0.577} \pm 0.043$ \\
 \hline
\multirow[c]{3}{*}{0.24m} & Chamfer $\boldsymbol{\downarrow}$ & $0.293 \pm 0.044$ & $0.272 \pm 0.055$ & $\mathbf{0.186} \pm 0.034$ \\
 & Precision $\boldsymbol{\uparrow}$ & $0.251 \pm 0.049$ & $0.225 \pm 0.039$ & $\mathbf{0.440} \pm 0.116$ \\
 & Recall $\boldsymbol{\uparrow}$ & $0.277 \pm 0.103$ & $0.290 \pm 0.076$ & $\mathbf{0.483} \pm 0.130$ \\
\bottomrule
\end{tabular}
}
\end{table}

\begin{table}[hb!]
\caption{Quantitative metrics for the seastar mesh.}
\label{table:seastar}
\resizebox{.9\linewidth}{!}{
\begin{tabular}{l|c|c|c|c}
\toprule
 &  & NeuS & NeuSIS & AONeuS \\
\midrule
\multirow[c]{3}{*}{1.2m} & Chamfer $\boldsymbol{\downarrow}$ & $0.108 \pm 0.020$ & $0.177 \pm 0.010$ & $\mathbf{0.088} \pm 0.022$ \\
 & Precision $\boldsymbol{\uparrow}$ & $0.585 \pm 0.070$ & $0.335 \pm 0.021$ & $\mathbf{0.714} \pm 0.106$ \\
 & Recall $\boldsymbol{\uparrow}$ & $0.722 \pm 0.061$ & $\mathbf{0.894} \pm 0.031$ & $0.886 \pm 0.026$ \\
 \hline
\multirow[c]{3}{*}{0.96m} & Chamfer $\boldsymbol{\downarrow}$ & $0.122 \pm 0.030$ & $0.170 \pm 0.014$ & $\mathbf{0.089} \pm 0.016$ \\
 & Precision $\boldsymbol{\uparrow}$ & $0.539 \pm 0.117$ & $0.352 \pm 0.032$ & $\mathbf{0.720} \pm 0.063$ \\
 & Recall $\boldsymbol{\uparrow}$ & $0.689 \pm 0.122$ & $\mathbf{0.848} \pm 0.022$ & $0.829 \pm 0.033$ \\
 \hline
\multirow[c]{3}{*}{0.72m} & Chamfer $\boldsymbol{\downarrow}$ & $0.159 \pm 0.035$ & $0.171 \pm 0.010$ & $\mathbf{0.126} \pm 0.035$ \\
 & Precision $\boldsymbol{\uparrow}$ & $0.435 \pm 0.089$ & $0.352 \pm 0.033$ & $\mathbf{0.571} \pm 0.118$ \\
 & Recall $\boldsymbol{\uparrow}$ & $0.550 \pm 0.127$ & $\mathbf{0.791} \pm 0.026$ & $0.764 \pm 0.092$ \\
 \hline
\multirow[c]{3}{*}{0.48m} & Chamfer $\boldsymbol{\downarrow}$ & $0.255 \pm 0.041$ & $0.170 \pm 0.006$ & $\mathbf{0.143} \pm 0.046$ \\
 & Precision $\boldsymbol{\uparrow}$ & $0.175 \pm 0.059$ & $0.372 \pm 0.019$ & $\mathbf{0.486} \pm 0.121$ \\
 & Recall $\boldsymbol{\uparrow}$ & $0.201 \pm 0.081$ & $\mathbf{0.742} \pm 0.039$ & $0.657 \pm 0.080$ \\
 \hline
\multirow[c]{3}{*}{0.24m} & Chamfer $\boldsymbol{\downarrow}$ & $0.548 \pm 0.144$ & $\mathbf{0.191} \pm 0.006$ & $0.196 \pm 0.033$ \\
 & Precision $\boldsymbol{\uparrow}$ & $0.067 \pm 0.038$ & $0.344 \pm 0.023$ & $\mathbf{0.377} \pm 0.088$ \\
 & Recall $\boldsymbol{\uparrow}$ & $0.071 \pm 0.045$ & $\mathbf{0.627} \pm 0.069$ & $0.516 \pm 0.072$ \\
\bottomrule
\end{tabular}
}
\end{table}

\begin{table}[hb!]
\caption{Quantitative metrics for the shell mesh.}
\label{table:shell}
\resizebox{.9\linewidth}{!}{
\begin{tabular}{l|c|c|c|c}
\toprule
 &  & NeuS & NeuSIS & AONeuS \\
\midrule
\multirow[c]{3}{*}{1.2m} & Chamfer $\boldsymbol{\downarrow}$ & $\mathbf{0.063} \pm 0.002$ & $0.077 \pm 0.010$ & $0.066 \pm 0.006$ \\
 & Precision $\boldsymbol{\uparrow}$ & $0.847 \pm 0.011$ & $0.754 \pm 0.066$ & $\mathbf{0.858} \pm 0.024$ \\
 & Recall $\boldsymbol{\uparrow}$ & $\mathbf{0.941} \pm 0.018$ & $0.814 \pm 0.035$ & $0.844 \pm 0.038$ \\
 \hline
\multirow[c]{3}{*}{0.96m} & Chamfer $\boldsymbol{\downarrow}$ & $\mathbf{0.068} \pm 0.005$ & $0.078 \pm 0.012$ & $0.078 \pm 0.006$ \\
 & Precision $\boldsymbol{\uparrow}$ & $\mathbf{0.833} \pm 0.023$ & $0.756 \pm 0.072$ & $0.816 \pm 0.022$ \\
 & Recall $\boldsymbol{\uparrow}$ & $\mathbf{0.929} \pm 0.020$ & $0.803 \pm 0.043$ & $0.794 \pm 0.031$ \\
 \hline
\multirow[c]{3}{*}{0.72m} & Chamfer $\boldsymbol{\downarrow}$ & $\mathbf{0.078} \pm 0.006$ & $0.091 \pm 0.016$ & $0.090 \pm 0.006$ \\
 & Precision $\boldsymbol{\uparrow}$ & $0.769 \pm 0.035$ & $0.747 \pm 0.086$ & $\mathbf{0.774} \pm 0.029$ \\
 & Recall $\boldsymbol{\uparrow}$ & $\mathbf{0.876} \pm 0.044$ & $0.746 \pm 0.075$ & $0.730 \pm 0.021$ \\
 \hline
\multirow[c]{3}{*}{0.48m} & Chamfer $\boldsymbol{\downarrow}$ & $0.107 \pm 0.014$ & $0.107 \pm 0.017$ & $\mathbf{0.098} \pm 0.009$ \\
 & Precision $\boldsymbol{\uparrow}$ & $0.620 \pm 0.064$ & $\mathbf{0.791} \pm 0.077$ & $0.714 \pm 0.031$ \\
 & Recall $\boldsymbol{\uparrow}$ & $0.690 \pm 0.084$ & $0.676 \pm 0.078$ & $\mathbf{0.691} \pm 0.037$ \\
 \hline
\multirow[c]{3}{*}{0.24m} & Chamfer $\boldsymbol{\downarrow}$ & $0.199 \pm 0.029$ & $0.168 \pm 0.008$ & $\mathbf{0.109} \pm 0.010$ \\
 & Precision $\boldsymbol{\uparrow}$ & $0.309 \pm 0.028$ & $\mathbf{0.832} \pm 0.041$ & $0.667 \pm 0.048$ \\
 & Recall $\boldsymbol{\uparrow}$ & $0.302 \pm 0.029$ & $0.465 \pm 0.032$ & $\mathbf{0.640} \pm 0.047$ \\
\bottomrule
\end{tabular}
}
\end{table}

\begin{table*}
    \caption{For the hardware reconstruction of "H" object, we report the mean and standard deviation of the Chamfer L1 distance, precision, and recall (with a threshold of \qty{0.05}{m}) compared to the ground truth (obtained from a laser scan of the real structure) for various reconstruction techniques. We computed the standard deviation over 6 trials. 
    For all methods, we compute the metrics for the intermediate reconstructions throughout training and report the best results. 
    }
    \label{table:realtableH}
\vspace{2mm}
    \begin{center}
    \resizebox{0.9\linewidth}{!}{
        \begin{tabular}{c|c|c|c|c|c|c|c}
        \toprule 
        \multicolumn{4}{c|}{} &  \multicolumn{2}{c|}{\makecell{Sonar dataset 1 \\ $14^\circ$ elevation angle} } & \multicolumn{2}{c}{\makecell{Sonar dataset 2 \\ $28^\circ$ elevation angle} }\\
            \hline
            Baseline & Metric &NeuS  & \makecell{Kim et al.\\ (2019)}&  \makecell{NeuSIS \\ ($14^\circ$)} &  \makecell{AONeuS \\($14^\circ$)} & \makecell{NeuSIS \\ ($28^\circ$)} &  \makecell{AONeuS \\ ($28^\circ$)}  \\
            \hline
    1.2m & Chamfer L1 $\boldsymbol{^\downarrow}$ & $\mathbf{0.092}\pm0.015$ & 0.177 & $0.159\pm0.032$  & $\textbf{0.092}\pm 0.007$ & $0.151\pm0.014$ &   $0.101\pm0.007$\\
     &Precision $\boldsymbol{^\uparrow}$& $\mathbf{0.693}\pm0.066$  &0.336  & $0.553\pm0.074$  & $0.661\pm0.040$ &   $0.513\pm0.061$&  $0.569\pm0.066$ \\
     &Recall $\boldsymbol{^\uparrow}$   & $0.679\pm0.098$ & 0.387 & $0.417\pm0.056$ &  $\textbf{0.708}\pm0.046$ &  $0.583\pm0.068$& $0.702\pm0.049$ \\
        \hline
    0.96m & Chamfer L1 $\boldsymbol{^\downarrow}$& $0.107\pm0.013$   &  0.182&  $0.158 \pm0.023$ & $\mathbf{0.088}\pm0.007$ &  $0.154\pm0.012$ & $0.093\pm0.004$ \\
     &Precision $\boldsymbol{^\uparrow}$&$0.661\pm0.048$ & 0.318& $0.560\pm0.068$ & $\mathbf{0.687}\pm0.019$ & $0.500\pm0.026$ & $0.602\pm0.045$\\
     &Recall $\boldsymbol{^\uparrow}$&$0.563\pm0.084$   & 0.345& $0.420\pm0.058$ &   $0.722\pm0.039$ & $0.562\pm0.042$ & $\mathbf{0.723}\pm0.034$\\
        \hline
    0.72m & Chamfer L1 $\boldsymbol{^\downarrow}$&$0.127\pm0.013$  &  0.178 & $0.167\pm0.029$ & $0.095\pm0.008$ &  $0.154\pm 0.013$ & $\textbf{0.088}\pm 0.003$ \\
     &Precision $\boldsymbol{^\uparrow}$& $0.651\pm0.047$ & 0.368 &  $0.562\pm0.077$ & $\textbf{0.667}\pm0.041$ &  $0.502\pm 0.054$ & $0.636 \pm 0.025$ \\
     &Recall $\boldsymbol{^\uparrow}$& $0.500\pm0.062$  & 0.396& $0.402\pm0.057$ & $0.688\pm0.008$ & $0.586\pm0.043$ &  $\textbf{0.748}\pm0.026$\\
        \hline
    0.48m & Chamfer L1 $\boldsymbol{^\downarrow}$& $0.150\pm0.022$ & 0.179 &  $0.170\pm0.028$ & $0.089\pm0.005$ & $0.143\pm0.007$ &  $ \textbf{0.086}\pm0.001$ \\
     &Precision $\boldsymbol{^\uparrow}$&$0.626\pm0.055$ & 0.324  & $0.543\pm0.045$ & $\mathbf{0.668}\pm0.006$ &  $0.547\pm0.021$& $0.636\pm0.022$ \\
     &Recall $\boldsymbol{^\uparrow}$&$0.415\pm0.022$  & 0.218 &  $0.395\pm0.066$ & $0.726\pm0.021$ & $0.605\pm0.021$ & $\textbf{0.757}\pm0.019$\\
        \hline
    0.24m & Chamfer L1 $\boldsymbol{^\downarrow}$& $0.167\pm0.012$  & 0.198 & $0.163\pm0.019$  & $\mathbf{0.085}\pm0.009$ & $0.148\pm0.017$ &  $0.108\pm0.005$\\
     &Precision $\boldsymbol{^\uparrow}$& $0.580\pm0.031$ & 0.305 & $0.551\pm0.058$ & $\mathbf{0.706}\pm0.063$& $0.481 \pm 0.056$ & $0.538\pm0.022$ \\
    &Recall $\boldsymbol{^\uparrow}$& $0.363\pm0.056$ & 0.140 &   $0.385\pm 0.039$  & $\mathbf{0.758}\pm0.041$ & $0.529 \pm 0.047$  &  $0.695\pm0.036$ \\
        \hline
        \end{tabular}
    }
    \end{center}

\end{table*}

%% file: sample-acmtog-SIGGRAPH-submission.bbl

\begin{thebibliography}{67}


\ifx \showCODEN    \undefined \def \showCODEN     #1{\unskip}     \fi
\ifx \showDOI      \undefined \def \showDOI       #1{#1}\fi
\ifx \showISBNx    \undefined \def \showISBNx     #1{\unskip}     \fi
\ifx \showISBNxiii \undefined \def \showISBNxiii  #1{\unskip}     \fi
\ifx \showISSN     \undefined \def \showISSN      #1{\unskip}     \fi
\ifx \showLCCN     \undefined \def \showLCCN      #1{\unskip}     \fi
\ifx \shownote     \undefined \def \shownote      #1{#1}          \fi
\ifx \showarticletitle \undefined \def \showarticletitle #1{#1}   \fi
\ifx \showURL      \undefined \def \showURL       {\relax}        \fi
\providecommand\bibfield[2]{#2}
\providecommand\bibinfo[2]{#2}
\providecommand\natexlab[1]{#1}
\providecommand\showeprint[2][]{arXiv:#2}

\bibitem[Albiez et~al\mbox{.}(2015)]%
        {albiez2015flatfish}
\bibfield{author}{\bibinfo{person}{Jan Albiez}, \bibinfo{person}{Sylvain Joyeux}, \bibinfo{person}{Christopher Gaudig}, \bibinfo{person}{Jens Hilljegerdes}, \bibinfo{person}{Sven Kroffke}, \bibinfo{person}{Christian Schoo}, \bibinfo{person}{Sascha Arnold}, \bibinfo{person}{Geovane Mimoso}, \bibinfo{person}{Pedro Alcantara}, \bibinfo{person}{Rafael Saback}, {et~al\mbox{.}}} \bibinfo{year}{2015}\natexlab{}.
\newblock \showarticletitle{Flatfish-a compact subsea-resident inspection auv}. In \bibinfo{booktitle}{\emph{OCEANS 2015-MTS/IEEE Washington}}. IEEE, \bibinfo{publisher}{IEEE}, \bibinfo{address}{201 Waterfront Street National Harbor, Maryland 20745 USA}, \bibinfo{pages}{1--8}.
\newblock


\bibitem[Arnold and Wehbe(2022)]%
        {arnold2022spatial}
\bibfield{author}{\bibinfo{person}{Sascha Arnold} {and} \bibinfo{person}{Bilal Wehbe}.} \bibinfo{year}{2022}\natexlab{}.
\newblock \showarticletitle{Spatial Acoustic Projection for 3D Imaging Sonar Reconstruction}. In \bibinfo{booktitle}{\emph{2022 International Conference on Robotics and Automation (ICRA)}}. \bibinfo{publisher}{IEEE}, \bibinfo{address}{Philadelphia}, \bibinfo{pages}{3054--3060}.
\newblock
\urldef\tempurl%
\url{https://doi.org/10.1109/ICRA46639.2022.9812277}
\showDOI{\tempurl}


\bibitem[Attal et~al\mbox{.}(2021)]%
        {attal2021torf}
\bibfield{author}{\bibinfo{person}{Benjamin Attal}, \bibinfo{person}{Eliot Laidlaw}, \bibinfo{person}{Aaron Gokaslan}, \bibinfo{person}{Changil Kim}, \bibinfo{person}{Christian Richardt}, \bibinfo{person}{James Tompkin}, {and} \bibinfo{person}{Matthew O'~Toole}.} \bibinfo{year}{2021}\natexlab{}.
\newblock \showarticletitle{{{T{\"o}RF}}: {{Time-of-Flight Radiance Fields}} for {{Dynamic Scene View Synthesis}}}. In \bibinfo{booktitle}{\emph{Advances in {{Neural Information Processing Systems}}}}, Vol.~\bibinfo{volume}{34}. \bibinfo{publisher}{Curran Associates, Inc.}, \bibinfo{address}{Virtual, Web}, \bibinfo{pages}{26289--26301}.
\newblock


\bibitem[Aykin and Negahdaripour(2015)]%
        {aykin20153}
\bibfield{author}{\bibinfo{person}{Murat~D Aykin} {and} \bibinfo{person}{Shahriar Negahdaripour}.} \bibinfo{year}{2015}\natexlab{}.
\newblock \showarticletitle{On 3-D target reconstruction from multiple 2-D forward-scan sonar views}. In \bibinfo{booktitle}{\emph{OCEANS 2015-Genova}}. IEEE, \bibinfo{publisher}{Institute of Electrical and Electronics Engineers}, \bibinfo{address}{Genova, Italy}, \bibinfo{pages}{1--10}.
\newblock


\bibitem[Aykin and Negahdaripour(2016a)]%
        {aykin2016three}
\bibfield{author}{\bibinfo{person}{Murat~D Aykin} {and} \bibinfo{person}{Shahriar Negahdaripour}.} \bibinfo{year}{2016}\natexlab{a}.
\newblock \showarticletitle{Three-dimensional target reconstruction from multiple 2-d forward-scan sonar views by space carving}.
\newblock \bibinfo{journal}{\emph{IEEE Journal of Oceanic Engineering}} \bibinfo{volume}{42}, \bibinfo{number}{3} (\bibinfo{year}{2016}), \bibinfo{pages}{574--589}.
\newblock


\bibitem[Aykin and Negahdaripour(2016b)]%
        {aykin2016modeling}
\bibfield{author}{\bibinfo{person}{Murat~D Aykin} {and} \bibinfo{person}{Shahriar~S Negahdaripour}.} \bibinfo{year}{2016}\natexlab{b}.
\newblock \showarticletitle{Modeling 2-D lens-based forward-scan sonar imagery for targets with diffuse reflectance}.
\newblock \bibinfo{journal}{\emph{IEEE journal of oceanic engineering}} \bibinfo{volume}{41}, \bibinfo{number}{3} (\bibinfo{year}{2016}), \bibinfo{pages}{569--582}.
\newblock


\bibitem[Babaee and Negahdaripour(2015)]%
        {babaee20153}
\bibfield{author}{\bibinfo{person}{Mohammadreza Babaee} {and} \bibinfo{person}{Shahriar Negahdaripour}.} \bibinfo{year}{2015}\natexlab{}.
\newblock \showarticletitle{3-D object modeling from 2-D occluding contour correspondences by opti-acoustic stereo imaging}.
\newblock \bibinfo{journal}{\emph{Computer Vision and Image Understanding}}  \bibinfo{volume}{132} (\bibinfo{year}{2015}), \bibinfo{pages}{56--74}.
\newblock


\bibitem[Bernardini et~al\mbox{.}(1999)]%
        {bernardini1999ball}
\bibfield{author}{\bibinfo{person}{Fausto Bernardini}, \bibinfo{person}{Joshua Mittleman}, \bibinfo{person}{Holly Rushmeier}, \bibinfo{person}{Cl{\'a}udio Silva}, {and} \bibinfo{person}{Gabriel Taubin}.} \bibinfo{year}{1999}\natexlab{}.
\newblock \showarticletitle{The ball-pivoting algorithm for surface reconstruction}.
\newblock \bibinfo{journal}{\emph{IEEE transactions on visualization and computer graphics}} \bibinfo{volume}{5}, \bibinfo{number}{4} (\bibinfo{year}{1999}), \bibinfo{pages}{349--359}.
\newblock


\bibitem[Bijelic et~al\mbox{.}(2020)]%
        {bijelic2020seeing}
\bibfield{author}{\bibinfo{person}{Mario Bijelic}, \bibinfo{person}{Tobias Gruber}, \bibinfo{person}{Fahim Mannan}, \bibinfo{person}{Florian Kraus}, \bibinfo{person}{Werner Ritter}, \bibinfo{person}{Klaus Dietmayer}, {and} \bibinfo{person}{Felix Heide}.} \bibinfo{year}{2020}\natexlab{}.
\newblock \showarticletitle{Seeing through fog without seeing fog: Deep multimodal sensor fusion in unseen adverse weather}. In \bibinfo{booktitle}{\emph{Proceedings of the IEEE/CVF Conference on Computer Vision and Pattern Recognition}}. \bibinfo{publisher}{Institute of Electrical and Electronics Engineers}, \bibinfo{address}{Virtual, Web}, \bibinfo{pages}{11682--11692}.
\newblock


\bibitem[Cardaillac and Ludvigsen(2023)]%
        {cardaillac2023camera}
\bibfield{author}{\bibinfo{person}{Alexandre Cardaillac} {and} \bibinfo{person}{Martin Ludvigsen}.} \bibinfo{year}{2023}\natexlab{}.
\newblock \showarticletitle{Camera-{{Sonar Combination}} for {{Improved Underwater Localization}} and {{Mapping}}}.
\newblock \bibinfo{journal}{\emph{IEEE Access}}  \bibinfo{volume}{11} (\bibinfo{year}{2023}), \bibinfo{pages}{123070--123079}.
\newblock
\showISSN{2169-3536}
\urldef\tempurl%
\url{https://doi.org/10.1109/ACCESS.2023.3329834}
\showDOI{\tempurl}


\bibitem[Carlson et~al\mbox{.}(2023)]%
        {10081483}
\bibfield{author}{\bibinfo{person}{Alexandra Carlson}, \bibinfo{person}{Manikandasriram~S. Ramanagopal}, \bibinfo{person}{Nathan Tseng}, \bibinfo{person}{Matthew Johnson-Roberson}, \bibinfo{person}{Ram Vasudevan}, {and} \bibinfo{person}{Katherine~A. Skinner}.} \bibinfo{year}{2023}\natexlab{}.
\newblock \showarticletitle{CLONeR: Camera-Lidar Fusion for Occupancy Grid-Aided Neural Representations}.
\newblock \bibinfo{journal}{\emph{IEEE Robotics and Automation Letters}} \bibinfo{volume}{8}, \bibinfo{number}{5} (\bibinfo{year}{2023}), \bibinfo{pages}{2812--2819}.
\newblock
\urldef\tempurl%
\url{https://doi.org/10.1109/LRA.2023.3262139}
\showDOI{\tempurl}


\bibitem[Chen et~al\mbox{.}(2024)]%
        {Chen2024dehazenerf}
\bibfield{author}{\bibinfo{person}{W. Chen}, \bibinfo{person}{W. Yifan}, \bibinfo{person}{S. Kuo}, {and} \bibinfo{person}{G. Wetzstein}.} \bibinfo{year}{2024}\natexlab{}.
\newblock \showarticletitle{DehazeNeRF: Multiple Image Haze Removal and 3D Shape Reconstruction using Neural Radiance Fields}. In \bibinfo{booktitle}{\emph{3DV}}. \bibinfo{publisher}{IEEE Computer Society}, \bibinfo{address}{Davos, Switzerland}, \bibinfo{pages}{no--pagination}.
\newblock


\bibitem[Chugunov et~al\mbox{.}(2024)]%
        {chugunov2024neural}
\bibfield{author}{\bibinfo{person}{Ilya Chugunov}, \bibinfo{person}{David Shustin}, \bibinfo{person}{Ruyu Yan}, \bibinfo{person}{Chenyang Lei}, {and} \bibinfo{person}{Felix Heide}.} \bibinfo{year}{2024}\natexlab{}.
\newblock \showarticletitle{Neural Spline Fields for Burst Image Fusion and Layer Separation}.
\newblock \bibinfo{journal}{\emph{CVPR}} (\bibinfo{year}{2024}).
\newblock


\bibitem[Chugunov et~al\mbox{.}(2023)]%
        {chugunov2023shakes}
\bibfield{author}{\bibinfo{person}{Ilya Chugunov}, \bibinfo{person}{Yuxuan Zhang}, {and} \bibinfo{person}{Felix Heide}.} \bibinfo{year}{2023}\natexlab{}.
\newblock \showarticletitle{Shakes on a Plane: Unsupervised Depth Estimation from Unstabilized Photography}. In \bibinfo{booktitle}{\emph{Proceedings of the IEEE/CVF Conference on Computer Vision and Pattern Recognition (CVPR)}}. \bibinfo{pages}{13240--13251}.
\newblock


\bibitem[Chugunov et~al\mbox{.}(2022)]%
        {chugunov2022implicit}
\bibfield{author}{\bibinfo{person}{Ilya Chugunov}, \bibinfo{person}{Yuxuan Zhang}, \bibinfo{person}{Zhihao Xia}, \bibinfo{person}{Xuaner Zhang}, \bibinfo{person}{Jiawen Chen}, {and} \bibinfo{person}{Felix Heide}.} \bibinfo{year}{2022}\natexlab{}.
\newblock \showarticletitle{The Implicit Values of A Good Hand Shake: Handheld Multi-Frame Neural Depth Refinement}. In \bibinfo{booktitle}{\emph{Proceedings of the IEEE/CVF Conference on Computer Vision and Pattern Recognition}}. \bibinfo{pages}{2852--2862}.
\newblock


\bibitem[Community(2022)]%
        {blender}
\bibfield{author}{\bibinfo{person}{Blender~Online Community}.} \bibinfo{year}{2022}\natexlab{}.
\newblock \bibinfo{booktitle}{\emph{Blender - a 3D modelling and rendering package}}.
\newblock Blender Foundation, Stichting Blender Foundation, Amsterdam.
\newblock
\urldef\tempurl%
\url{http://www.blender.org}
\showURL{%
\tempurl}


\bibitem[Dave et~al\mbox{.}(2022)]%
        {Dave2022}
\bibfield{author}{\bibinfo{person}{Akshat Dave}, \bibinfo{person}{Yongyi Zhao}, {and} \bibinfo{person}{Ashok Veeraraghavan}.} \bibinfo{year}{2022}\natexlab{}.
\newblock \bibinfo{booktitle}{\emph{PANDORA: Polarization-Aided Neural Decomposition of Radiance}}.
\newblock \bibinfo{publisher}{Springer Nature Switzerland}, \bibinfo{address}{Tel Aviv, Israel}, \bibinfo{pages}{538–556}.
\newblock
\showISBNx{9783031200717}
\showISSN{1611-3349}
\urldef\tempurl%
\url{https://doi.org/10.1007/978-3-031-20071-7_32}
\showDOI{\tempurl}


\bibitem[DeBortoli et~al\mbox{.}(2019)]%
        {debortoli2019elevatenet}
\bibfield{author}{\bibinfo{person}{Robert DeBortoli}, \bibinfo{person}{Fuxin Li}, {and} \bibinfo{person}{Geoffrey~A Hollinger}.} \bibinfo{year}{2019}\natexlab{}.
\newblock \showarticletitle{Elevatenet: A convolutional neural network for estimating the missing dimension in 2d underwater sonar images}. In \bibinfo{booktitle}{\emph{2019 IEEE/RSJ International Conference on Intelligent Robots and Systems (IROS)}}. IEEE, \bibinfo{publisher}{Institute for Electrical and Electronics Engineers}, \bibinfo{address}{Macau, China}, \bibinfo{pages}{8040--8047}.
\newblock


\bibitem[Ferreira et~al\mbox{.}(2016)]%
        {ferreira2016underwater}
\bibfield{author}{\bibinfo{person}{Fausto Ferreira}, \bibinfo{person}{Diogo Machado}, \bibinfo{person}{Gabriele Ferri}, \bibinfo{person}{Samantha Dugelay}, {and} \bibinfo{person}{John Potter}.} \bibinfo{year}{2016}\natexlab{}.
\newblock \showarticletitle{Underwater Optical and Acoustic Imaging: {{A}} Time for Fusion? A Brief Overview of the State-of-the-Art}. In \bibinfo{booktitle}{\emph{{{OCEANS}} 2016 {{MTS}}/{{IEEE Monterey}}}}. \bibinfo{publisher}{Institute of Electrical and Electronics Engineers}, \bibinfo{address}{Monterey, USA}, \bibinfo{pages}{1--6}.
\newblock
\urldef\tempurl%
\url{https://doi.org/10.1109/OCEANS.2016.7761354}
\showDOI{\tempurl}


\bibitem[Goli et~al\mbox{.}(2023)]%
        {goli2023}
\bibfield{author}{\bibinfo{person}{Lily Goli}, \bibinfo{person}{Cody Reading}, \bibinfo{person}{Silvia Sellán}, \bibinfo{person}{Alec Jacobson}, {and} \bibinfo{person}{Andrea Tagliasacchi}.} \bibinfo{year}{2023}\natexlab{}.
\newblock \bibinfo{title}{Bayes' Rays: Uncertainty Quantification for Neural Radiance Fields}.
\newblock
\newblock
\showeprint[arxiv]{2309.03185}~[cs.CV]


\bibitem[Iscar et~al\mbox{.}(2017)]%
        {iscar2017multi}
\bibfield{author}{\bibinfo{person}{Eduardo Iscar}, \bibinfo{person}{Katherine~A Skinner}, {and} \bibinfo{person}{Matthew Johnson-Roberson}.} \bibinfo{year}{2017}\natexlab{}.
\newblock \showarticletitle{Multi-view 3D reconstruction in underwater environments: Evaluation and benchmark}. In \bibinfo{booktitle}{\emph{OCEANS 2017-Anchorage}}. IEEE, \bibinfo{publisher}{Institute of Electrical and Electronics Engineers}, \bibinfo{address}{Anchorage, Alaska}, \bibinfo{pages}{1--8}.
\newblock


\bibitem[Jaffe(2014)]%
        {jaffe2014underwater}
\bibfield{author}{\bibinfo{person}{Jules~S Jaffe}.} \bibinfo{year}{2014}\natexlab{}.
\newblock \showarticletitle{Underwater optical imaging: the past, the present, and the prospects}.
\newblock \bibinfo{journal}{\emph{IEEE Journal of Oceanic Engineering}} \bibinfo{volume}{40}, \bibinfo{number}{3} (\bibinfo{year}{2014}), \bibinfo{pages}{683--700}.
\newblock


\bibitem[Jiang et~al\mbox{.}(2023)]%
        {jiang2023fisherrf}
\bibfield{author}{\bibinfo{person}{Wen Jiang}, \bibinfo{person}{Boshu Lei}, {and} \bibinfo{person}{Kostas Daniilidis}.} \bibinfo{year}{2023}\natexlab{}.
\newblock \bibinfo{title}{FisherRF: Active View Selection and Uncertainty Quantification for Radiance Fields using Fisher Information}.
\newblock
\newblock
\showeprint[arxiv]{2311.17874}~[cs.CV]


\bibitem[Johnson-Roberson et~al\mbox{.}(2010)]%
        {johnson2010generation}
\bibfield{author}{\bibinfo{person}{Matthew Johnson-Roberson}, \bibinfo{person}{Oscar Pizarro}, \bibinfo{person}{Stefan~B Williams}, {and} \bibinfo{person}{Ian Mahon}.} \bibinfo{year}{2010}\natexlab{}.
\newblock \showarticletitle{Generation and visualization of large-scale three-dimensional reconstructions from underwater robotic surveys}.
\newblock \bibinfo{journal}{\emph{Journal of Field Robotics}} \bibinfo{volume}{27}, \bibinfo{number}{1} (\bibinfo{year}{2010}), \bibinfo{pages}{21--51}.
\newblock


\bibitem[Kim et~al\mbox{.}(2019)]%
        {kim3DReconstructionUnderwater2019}
\bibfield{author}{\bibinfo{person}{Jason Kim}, \bibinfo{person}{Meungsuk Lee}, \bibinfo{person}{Seokyong Song}, \bibinfo{person}{Byeongjin Kim}, {and} \bibinfo{person}{Son-Cheol Yu}.} \bibinfo{year}{2019}\natexlab{}.
\newblock \showarticletitle{3-{{D Reconstruction}} of {{Underwater Objects Using Image Sequences}} from {{Optical Camera}} and {{Imaging Sonar}}}. In \bibinfo{booktitle}{\emph{{{OCEANS}} 2019 {{MTS}}/{{IEEE SEATTLE}}}}. \bibinfo{publisher}{Institute of Electrical and Electronics Engineers}, \bibinfo{address}{Seattle, USA}, \bibinfo{pages}{1--6}.
\newblock
\showISSN{0197-7385}
\urldef\tempurl%
\url{https://doi.org/10.23919/OCEANS40490.2019.8962558}
\showDOI{\tempurl}


\bibitem[Kim et~al\mbox{.}(2023)]%
        {10.1145/3610548.3618172}
\bibfield{author}{\bibinfo{person}{Youngchan Kim}, \bibinfo{person}{Wonjoon Jin}, \bibinfo{person}{Sunghyun Cho}, {and} \bibinfo{person}{Seung-Hwan Baek}.} \bibinfo{year}{2023}\natexlab{}.
\newblock \showarticletitle{Neural Spectro-polarimetric Fields}. In \bibinfo{booktitle}{\emph{SIGGRAPH Asia 2023 Conference Papers}} (, Sydney, NSW, Australia,) \emph{(\bibinfo{series}{SA '23})}. \bibinfo{publisher}{Association for Computing Machinery}, \bibinfo{address}{New York, NY, USA}, Article \bibinfo{articleno}{109}, \bibinfo{numpages}{11}~pages.
\newblock
\showISBNx{9798400703157}
\urldef\tempurl%
\url{https://doi.org/10.1145/3610548.3618172}
\showDOI{\tempurl}


\bibitem[Kim et~al\mbox{.}(2009)]%
        {5457430}
\bibfield{author}{\bibinfo{person}{Young~Min Kim}, \bibinfo{person}{Christian Theobalt}, \bibinfo{person}{James Diebel}, \bibinfo{person}{Jana Kosecka}, \bibinfo{person}{Branislav Miscusik}, {and} \bibinfo{person}{Sebastian Thrun}.} \bibinfo{year}{2009}\natexlab{}.
\newblock \showarticletitle{Multi-view image and ToF sensor fusion for dense 3D reconstruction}. In \bibinfo{booktitle}{\emph{2009 IEEE 12th International Conference on Computer Vision Workshops, ICCV Workshops}}. \bibinfo{publisher}{Institute of Electrical and Electronics Engineers}, \bibinfo{address}{Kyoto, Japan}, \bibinfo{pages}{1542--1549}.
\newblock
\urldef\tempurl%
\url{https://doi.org/10.1109/ICCVW.2009.5457430}
\showDOI{\tempurl}


\bibitem[Langer and Hebert(1991)]%
        {langerBuildingQualitativeElevation1991}
\bibfield{author}{\bibinfo{person}{D. Langer} {and} \bibinfo{person}{M. Hebert}.} \bibinfo{year}{1991}\natexlab{}.
\newblock \showarticletitle{Building Qualitative Elevation Maps from Side Scan Sonar Data for Autonomous Underwater Navigation}. In \bibinfo{booktitle}{\emph{1991 {{IEEE International Conference}} on {{Robotics}} and {{Automation Proceedings}}}}. \bibinfo{publisher}{Institute of Electrical and Electronics Engineers}, \bibinfo{address}{Sacramento, USA}, \bibinfo{pages}{2478--2483 vol.3}.
\newblock
\urldef\tempurl%
\url{https://doi.org/10.1109/ROBOT.1991.131997}
\showDOI{\tempurl}


\bibitem[Lensgraf et~al\mbox{.}(2021)]%
        {lensgraf2021droplet}
\bibfield{author}{\bibinfo{person}{Samuel Lensgraf}, \bibinfo{person}{Amy Sniffen}, \bibinfo{person}{Zachary Zitzewitz}, \bibinfo{person}{Evan Honnold}, \bibinfo{person}{Jennifer Jain}, \bibinfo{person}{Weifu Wang}, \bibinfo{person}{Alberto Li}, {and} \bibinfo{person}{Devin Balkcom}.} \bibinfo{year}{2021}\natexlab{}.
\newblock \showarticletitle{Droplet: {{Towards Autonomous Underwater Assembly}} of {{Modular Structures}}}. In \bibinfo{booktitle}{\emph{Robotics: {{Science}} and {{Systems XVII}}}}. \bibinfo{publisher}{{Robotics: Science and Systems Foundation}}, \bibinfo{address}{Virtual, Web}, \bibinfo{pages}{no--pagination}.
\newblock
\showISBNx{978-0-9923747-7-8}
\urldef\tempurl%
\url{https://doi.org/10.15607/RSS.2021.XVII.054}
\showDOI{\tempurl}


\bibitem[Levy et~al\mbox{.}(2023)]%
        {levy2023seathru}
\bibfield{author}{\bibinfo{person}{Deborah Levy}, \bibinfo{person}{Amit Peleg}, \bibinfo{person}{Naama Pearl}, \bibinfo{person}{Dan Rosenbaum}, \bibinfo{person}{Derya Akkaynak}, \bibinfo{person}{Simon Korman}, {and} \bibinfo{person}{Tali Treibitz}.} \bibinfo{year}{2023}\natexlab{}.
\newblock \showarticletitle{SeaThru-NeRF: Neural Radiance Fields in Scattering Media}. In \bibinfo{booktitle}{\emph{Proceedings of the IEEE/CVF Conference on Computer Vision and Pattern Recognition}}. \bibinfo{publisher}{Institute of Electrical and Electronics Engineers}, \bibinfo{address}{Vancouver, Canada}, \bibinfo{pages}{56--65}.
\newblock


\bibitem[Lin et~al\mbox{.}(2023)]%
        {lin2023conditional}
\bibfield{author}{\bibinfo{person}{Tianxiang Lin}, \bibinfo{person}{Akshay Hinduja}, \bibinfo{person}{Mohamad Qadri}, {and} \bibinfo{person}{Michael Kaess}.} \bibinfo{year}{2023}\natexlab{}.
\newblock \showarticletitle{Conditional GANs for Sonar Image Filtering with Applications to Underwater Occupancy Mapping}. In \bibinfo{booktitle}{\emph{2023 IEEE International Conference on Robotics and Automation (ICRA)}}. IEEE, \bibinfo{publisher}{Institute of Electronics and Electrical Engineers}, \bibinfo{address}{London, England}, \bibinfo{pages}{1048--1054}.
\newblock


\bibitem[Lindell et~al\mbox{.}(2018)]%
        {Lindell:2018:3D}
\bibfield{author}{\bibinfo{person}{David~B. Lindell}, \bibinfo{person}{Matthew O'Toole}, {and} \bibinfo{person}{Gordon Wetzstein}.} \bibinfo{year}{2018}\natexlab{}.
\newblock \showarticletitle{Single-Photon {{3D}} Imaging with Deep Sensor Fusion}.
\newblock \bibinfo{journal}{\emph{ACM Transactions on Graphics}} \bibinfo{volume}{37}, \bibinfo{number}{4} (\bibinfo{date}{July} \bibinfo{year}{2018}), \bibinfo{pages}{113:1--113:12}.
\newblock
\showISSN{0730-0301}
\urldef\tempurl%
\url{https://doi.org/10.1145/3197517.3201316}
\showDOI{\tempurl}


\bibitem[Liu et~al\mbox{.}(2023b)]%
        {10190736}
\bibfield{author}{\bibinfo{person}{Afei Liu}, \bibinfo{person}{Shuanghui Zhang}, \bibinfo{person}{Chi Zhang}, \bibinfo{person}{Shuaifeng Zhi}, {and} \bibinfo{person}{Xiang Li}.} \bibinfo{year}{2023}\natexlab{b}.
\newblock \showarticletitle{RaNeRF: Neural 3-D Reconstruction of Space Targets From ISAR Image Sequences}.
\newblock \bibinfo{journal}{\emph{IEEE Transactions on Geoscience and Remote Sensing}}  \bibinfo{volume}{61} (\bibinfo{year}{2023}), \bibinfo{pages}{1--15}.
\newblock
\urldef\tempurl%
\url{https://doi.org/10.1109/TGRS.2023.3298067}
\showDOI{\tempurl}


\bibitem[Liu et~al\mbox{.}(2023a)]%
        {liu2023deep}
\bibfield{author}{\bibinfo{person}{Haowen Liu}, \bibinfo{person}{Monika Roznere}, {and} \bibinfo{person}{Alberto~Quattrini Li}.} \bibinfo{year}{2023}\natexlab{a}.
\newblock \showarticletitle{Deep Underwater Monocular Depth Estimation with Single-Beam Echosounder}. In \bibinfo{booktitle}{\emph{2023 IEEE International Conference on Robotics and Automation (ICRA)}}. IEEE, \bibinfo{publisher}{Institute of Electrical and Electronics Engineers}, \bibinfo{address}{London, England}, \bibinfo{pages}{1090--1097}.
\newblock


\bibitem[Long et~al\mbox{.}(2022)]%
        {long2022sparseneus}
\bibfield{author}{\bibinfo{person}{Xiaoxiao Long}, \bibinfo{person}{Cheng Lin}, \bibinfo{person}{Peng Wang}, \bibinfo{person}{Taku Komura}, {and} \bibinfo{person}{Wenping Wang}.} \bibinfo{year}{2022}\natexlab{}.
\newblock \showarticletitle{{{SparseNeuS}}: {{Fast Generalizable Neural Surface Reconstruction}} from {{Sparse Views}}}.
\newblock In \bibinfo{booktitle}{\emph{Computer {{Vision}} -- {{ECCV}} 2022}}, \bibfield{editor}{\bibinfo{person}{Shai Avidan}, \bibinfo{person}{Gabriel Brostow}, \bibinfo{person}{Moustapha Ciss{\'e}}, \bibinfo{person}{Giovanni~Maria Farinella}, {and} \bibinfo{person}{Tal Hassner}} (Eds.). Vol.~\bibinfo{volume}{13692}. \bibinfo{publisher}{Springer Nature Switzerland}, \bibinfo{address}{Cham}, \bibinfo{pages}{210--227}.
\newblock
\showISBNx{978-3-031-19823-6 978-3-031-19824-3}
\urldef\tempurl%
\url{https://doi.org/10.1007/978-3-031-19824-3_13}
\showDOI{\tempurl}


\bibitem[Malik et~al\mbox{.}(2023)]%
        {malik2023transient}
\bibfield{author}{\bibinfo{person}{Anagh Malik}, \bibinfo{person}{Parsa Mirdehghan}, \bibinfo{person}{Sotiris Nousias}, \bibinfo{person}{Kyros Kutulakos}, {and} \bibinfo{person}{David Lindell}.} \bibinfo{year}{2023}\natexlab{}.
\newblock \showarticletitle{Transient {{Neural Radiance Fields}} for {{Lidar View Synthesis}} and {{3D Reconstruction}}}. In \bibinfo{booktitle}{\emph{Advances in {{Neural Information Processing Systems}}}}, Vol.~\bibinfo{volume}{36}. \bibinfo{publisher}{Curran Associates, Inc.}, \bibinfo{address}{New Orleans, USA}, \bibinfo{pages}{71569--71581}.
\newblock


\bibitem[Menna et~al\mbox{.}(2018)]%
        {menna2018state}
\bibfield{author}{\bibinfo{person}{Fabio Menna}, \bibinfo{person}{Panagiotis Agrafiotis}, {and} \bibinfo{person}{Andreas Georgopoulos}.} \bibinfo{year}{2018}\natexlab{}.
\newblock \showarticletitle{State of the art and applications in archaeological underwater 3D recording and mapping}.
\newblock \bibinfo{journal}{\emph{Journal of Cultural Heritage}}  \bibinfo{volume}{33} (\bibinfo{year}{2018}), \bibinfo{pages}{231--248}.
\newblock


\bibitem[Mildenhall et~al\mbox{.}(2020)]%
        {mildenhall2020nerf}
\bibfield{author}{\bibinfo{person}{Ben Mildenhall}, \bibinfo{person}{Pratul~P. Srinivasan}, \bibinfo{person}{Matthew Tancik}, \bibinfo{person}{Jonathan~T. Barron}, \bibinfo{person}{Ravi Ramamoorthi}, {and} \bibinfo{person}{Ren Ng}.} \bibinfo{year}{2020}\natexlab{}.
\newblock \showarticletitle{{{NeRF}}: {{Representing Scenes}} as {{Neural Radiance Fields}} for {{View Synthesis}}}. In \bibinfo{booktitle}{\emph{Computer {{Vision}} -- {{ECCV}} 2020: 16th {{European Conference}}, {{Glasgow}}, {{UK}}, {{August}} 23--28, 2020, {{Proceedings}}, {{Part I}}}}. \bibinfo{publisher}{Springer-Verlag}, \bibinfo{address}{Berlin, Heidelberg}, \bibinfo{pages}{405--421}.
\newblock
\showISBNx{978-3-030-58451-1}
\urldef\tempurl%
\url{https://doi.org/10.1007/978-3-030-58452-8_24}
\showDOI{\tempurl}


\bibitem[Mur-Artal et~al\mbox{.}(2015)]%
        {mur2015orb}
\bibfield{author}{\bibinfo{person}{Raul Mur-Artal}, \bibinfo{person}{Jose Maria~Martinez Montiel}, {and} \bibinfo{person}{Juan~D Tardos}.} \bibinfo{year}{2015}\natexlab{}.
\newblock \showarticletitle{ORB-SLAM: a versatile and accurate monocular SLAM system}.
\newblock \bibinfo{journal}{\emph{IEEE transactions on robotics}} \bibinfo{volume}{31}, \bibinfo{number}{5} (\bibinfo{year}{2015}), \bibinfo{pages}{1147--1163}.
\newblock


\bibitem[Nayar et~al\mbox{.}(1991)]%
        {nayarSurfaceReflectionPhysical1991}
\bibfield{author}{\bibinfo{person}{S.K. Nayar}, \bibinfo{person}{K. Ikeuchi}, {and} \bibinfo{person}{T. Kanade}.} \bibinfo{year}{1991}\natexlab{}.
\newblock \showarticletitle{Surface Reflection: Physical and Geometrical Perspectives}.
\newblock \bibinfo{journal}{\emph{IEEE Transactions on Pattern Analysis and Machine Intelligence}} \bibinfo{volume}{13}, \bibinfo{number}{7} (\bibinfo{date}{July} \bibinfo{year}{1991}), \bibinfo{pages}{611--634}.
\newblock
\showISSN{1939-3539}
\urldef\tempurl%
\url{https://doi.org/10.1109/34.85654}
\showDOI{\tempurl}


\bibitem[Negahdaripour(2018)]%
        {negahdaripour2018application}
\bibfield{author}{\bibinfo{person}{Shahriar Negahdaripour}.} \bibinfo{year}{2018}\natexlab{}.
\newblock \showarticletitle{Application of forward-scan sonar stereo for 3-D scene reconstruction}.
\newblock \bibinfo{journal}{\emph{IEEE journal of oceanic engineering}} \bibinfo{volume}{45}, \bibinfo{number}{2} (\bibinfo{year}{2018}), \bibinfo{pages}{547--562}.
\newblock


\bibitem[Negahdaripour et~al\mbox{.}(2017)]%
        {negahdaripour2017refining}
\bibfield{author}{\bibinfo{person}{Shahriar Negahdaripour}, \bibinfo{person}{Victor~M Milenkovic}, \bibinfo{person}{Nikan Salarieh}, {and} \bibinfo{person}{Mahsa Mirzargar}.} \bibinfo{year}{2017}\natexlab{}.
\newblock \showarticletitle{Refining 3-D object models constructed from multiple FS sonar images by space carving}. In \bibinfo{booktitle}{\emph{OCEANS 2017-Anchorage}}. IEEE, \bibinfo{publisher}{Institute of Electrical and Electronics Engineers}, \bibinfo{address}{Anchorage, USA}, \bibinfo{pages}{1--9}.
\newblock


\bibitem[Negahdaripour et~al\mbox{.}(2009)]%
        {negahdaripour2009opti}
\bibfield{author}{\bibinfo{person}{Shahriar Negahdaripour}, \bibinfo{person}{Hicham Sekkati}, {and} \bibinfo{person}{Hamed Pirsiavash}.} \bibinfo{year}{2009}\natexlab{}.
\newblock \showarticletitle{Opti-acoustic stereo imaging: On system calibration and 3-D target reconstruction}.
\newblock \bibinfo{journal}{\emph{IEEE Transactions on image processing}} \bibinfo{volume}{18}, \bibinfo{number}{6} (\bibinfo{year}{2009}), \bibinfo{pages}{1203--1214}.
\newblock


\bibitem[Nishimura et~al\mbox{.}(2020)]%
        {nishimura2020disambiguating}
\bibfield{author}{\bibinfo{person}{Mark Nishimura}, \bibinfo{person}{David~B Lindell}, \bibinfo{person}{Christopher Metzler}, {and} \bibinfo{person}{Gordon Wetzstein}.} \bibinfo{year}{2020}\natexlab{}.
\newblock \showarticletitle{Disambiguating monocular depth estimation with a single transient}. In \bibinfo{booktitle}{\emph{European Conference on Computer Vision}}. Springer, \bibinfo{publisher}{Springer}, \bibinfo{address}{Virtual, Web}, \bibinfo{pages}{139--155}.
\newblock


\bibitem[Oechsle et~al\mbox{.}(2021)]%
        {Oechsle2021ICCV}
\bibfield{author}{\bibinfo{person}{Michael Oechsle}, \bibinfo{person}{Songyou Peng}, {and} \bibinfo{person}{Andreas Geiger}.} \bibinfo{year}{2021}\natexlab{}.
\newblock \showarticletitle{{{UNISURF}}: {{Unifying Neural Implicit Surfaces}} and {{Radiance Fields}} for {{Multi-View Reconstruction}}}. In \bibinfo{booktitle}{\emph{Proceedings of the {{IEEE}}/{{CVF International Conference}} on {{Computer Vision}}}}. \bibinfo{publisher}{Institute of Electrical and Electronics Engineers}, \bibinfo{address}{Virtual, Web}, \bibinfo{pages}{5589--5599}.
\newblock


\bibitem[Poggi et~al\mbox{.}(2022)]%
        {poggi2022xnerf}
\bibfield{author}{\bibinfo{person}{M. Poggi}, \bibinfo{person}{P. Ramirez}, \bibinfo{person}{F. Tosi}, \bibinfo{person}{S. Salti}, \bibinfo{person}{S. Mattoccia}, {and} \bibinfo{person}{L. Stefano}.} \bibinfo{year}{2022}\natexlab{}.
\newblock \showarticletitle{Cross-Spectral Neural Radiance Fields}. In \bibinfo{booktitle}{\emph{2022 International Conference on 3D Vision (3DV)}}. \bibinfo{publisher}{IEEE Computer Society}, \bibinfo{address}{Los Alamitos, CA, USA}, \bibinfo{pages}{606--616}.
\newblock
\urldef\tempurl%
\url{https://doi.org/10.1109/3DV57658.2022.00071}
\showDOI{\tempurl}


\bibitem[Qadri et~al\mbox{.}(2023)]%
        {qadri2023neural}
\bibfield{author}{\bibinfo{person}{Mohamad Qadri}, \bibinfo{person}{Michael Kaess}, {and} \bibinfo{person}{Ioannis Gkioulekas}.} \bibinfo{year}{2023}\natexlab{}.
\newblock \showarticletitle{Neural implicit surface reconstruction using imaging sonar}. In \bibinfo{booktitle}{\emph{2023 IEEE International Conference on Robotics and Automation (ICRA)}}. IEEE, \bibinfo{publisher}{Institute of Electrical and Electronics Engineers}, \bibinfo{address}{London, USA}, \bibinfo{pages}{1040--1047}.
\newblock


\bibitem[Ramazzina et~al\mbox{.}(2023)]%
        {ramazzina2023scatternerf}
\bibfield{author}{\bibinfo{person}{Andrea Ramazzina}, \bibinfo{person}{Mario Bijelic}, \bibinfo{person}{Stefanie Walz}, \bibinfo{person}{Alessandro Sanvito}, \bibinfo{person}{Dominik Scheuble}, {and} \bibinfo{person}{Felix Heide}.} \bibinfo{year}{2023}\natexlab{}.
\newblock \showarticletitle{{{ScatterNeRF}}: {{Seeing Through Fog}} with {{Physically-Based Inverse Neural Rendering}}}. In \bibinfo{booktitle}{\emph{Proceedings of the {{IEEE}}/{{CVF International Conference}} on {{Computer Vision}}}}. \bibinfo{publisher}{Institute of Electrical and Electronics Engineers}, \bibinfo{address}{Paris, France}, \bibinfo{pages}{17957--17968}.
\newblock


\bibitem[Reed et~al\mbox{.}(2023)]%
        {reedNeuralVolumetricReconstruction2023}
\bibfield{author}{\bibinfo{person}{Albert Reed}, \bibinfo{person}{Juhyeon Kim}, \bibinfo{person}{Thomas Blanford}, \bibinfo{person}{Adithya Pediredla}, \bibinfo{person}{Daniel Brown}, {and} \bibinfo{person}{Suren Jayasuriya}.} \bibinfo{year}{2023}\natexlab{}.
\newblock \showarticletitle{Neural {{Volumetric Reconstruction}} for {{Coherent Synthetic Aperture Sonar}}}.
\newblock \bibinfo{journal}{\emph{ACM Transactions on Graphics}} \bibinfo{volume}{42}, \bibinfo{number}{4} (\bibinfo{date}{July} \bibinfo{year}{2023}), \bibinfo{pages}{113:1--113:20}.
\newblock
\showISSN{0730-0301}
\urldef\tempurl%
\url{https://doi.org/10.1145/3592141}
\showDOI{\tempurl}


\bibitem[Roznere et~al\mbox{.}(2023)]%
        {roznere20233}
\bibfield{author}{\bibinfo{person}{Monika Roznere}, \bibinfo{person}{Philippos Mordohai}, \bibinfo{person}{Ioannis Rekleitis}, {and} \bibinfo{person}{Alberto~Quattrini Li}.} \bibinfo{year}{2023}\natexlab{}.
\newblock \showarticletitle{3-D Reconstruction Using Monocular Camera and Lights: Multi-View Photometric Stereo for Non-Stationary Robots}. In \bibinfo{booktitle}{\emph{2023 IEEE International Conference on Robotics and Automation (ICRA)}}. IEEE, \bibinfo{publisher}{Institute of Electronics and Electrical Engineers}, \bibinfo{address}{London, England}, \bibinfo{pages}{1026--1032}.
\newblock


\bibitem[Schonberger and Frahm(2016)]%
        {schonberger2016structure}
\bibfield{author}{\bibinfo{person}{Johannes~L Schonberger} {and} \bibinfo{person}{Jan-Michael Frahm}.} \bibinfo{year}{2016}\natexlab{}.
\newblock \showarticletitle{Structure-from-motion revisited}. In \bibinfo{booktitle}{\emph{Proceedings of the IEEE conference on computer vision and pattern recognition}}. \bibinfo{publisher}{Institute of Electrical and Electronics Engineers}, \bibinfo{address}{Las Vegas, USA}, \bibinfo{pages}{4104--4113}.
\newblock


\bibitem[Sethuraman et~al\mbox{.}(2023)]%
        {sethuraman2023waternerf}
\bibfield{author}{\bibinfo{person}{Advaith~Venkatramanan Sethuraman}, \bibinfo{person}{Manikandasriram~Srinivasan Ramanagopal}, {and} \bibinfo{person}{Katherine~A Skinner}.} \bibinfo{year}{2023}\natexlab{}.
\newblock \showarticletitle{Waternerf: Neural radiance fields for underwater scenes}. In \bibinfo{booktitle}{\emph{OCEANS 2023-MTS/IEEE US Gulf Coast}}. IEEE, \bibinfo{publisher}{Institute of Electrical and Electronics Engineers}, \bibinfo{address}{Biloxi, USA}, \bibinfo{pages}{1--7}.
\newblock


\bibitem[Teixeira et~al\mbox{.}(2016)]%
        {teixeira2016underwater}
\bibfield{author}{\bibinfo{person}{Pedro~V Teixeira}, \bibinfo{person}{Michael Kaess}, \bibinfo{person}{Franz~S Hover}, {and} \bibinfo{person}{John~J Leonard}.} \bibinfo{year}{2016}\natexlab{}.
\newblock \showarticletitle{Underwater inspection using sonar-based volumetric submaps}. In \bibinfo{booktitle}{\emph{2016 IEEE/RSJ International Conference on Intelligent Robots and Systems (IROS)}}. IEEE, \bibinfo{publisher}{Institute of Electrical and Electronics Engineers}, \bibinfo{address}{Daejeon, South Korea}, \bibinfo{pages}{4288--4295}.
\newblock


\bibitem[Tinh and Khanh(2021)]%
        {tinh2021new}
\bibfield{author}{\bibinfo{person}{Nguyen~Dinh Tinh} {and} \bibinfo{person}{T~Dang Khanh}.} \bibinfo{year}{2021}\natexlab{}.
\newblock \showarticletitle{A new imaging geometry model for multi-receiver synthetic aperture sonar considering variation of the speed of sound in seawater}.
\newblock \bibinfo{journal}{\emph{IEIE Transactions on Smart Processing and Computing}} \bibinfo{volume}{10}, \bibinfo{number}{4} (\bibinfo{year}{2021}), \bibinfo{pages}{302--308}.
\newblock


\bibitem[Torrance and Sparrow(1967)]%
        {torranceTheoryOffSpecularReflection1967}
\bibfield{author}{\bibinfo{person}{K.~E. Torrance} {and} \bibinfo{person}{E.~M. Sparrow}.} \bibinfo{year}{1967}\natexlab{}.
\newblock \showarticletitle{Theory for {{Off-Specular Reflection From Roughened Surfaces}}*}.
\newblock \bibinfo{journal}{\emph{Journal of the Optical Society of America}} \bibinfo{volume}{57}, \bibinfo{number}{9} (\bibinfo{date}{Sept.} \bibinfo{year}{1967}), \bibinfo{pages}{1105}.
\newblock
\showISSN{0030-3941}
\urldef\tempurl%
\url{https://doi.org/10.1364/JOSA.57.001105}
\showDOI{\tempurl}


\bibitem[Wang et~al\mbox{.}(2019b)]%
        {wang2019underwater}
\bibfield{author}{\bibinfo{person}{Jinkun Wang}, \bibinfo{person}{Tixiao Shan}, {and} \bibinfo{person}{Brendan Englot}.} \bibinfo{year}{2019}\natexlab{b}.
\newblock \showarticletitle{Underwater terrain reconstruction from forward-looking sonar imagery}. In \bibinfo{booktitle}{\emph{2019 International Conference on Robotics and Automation (ICRA)}}. IEEE, \bibinfo{publisher}{Institute of Electrical and Electronics Engineers}, \bibinfo{address}{Montreal, Canada}, \bibinfo{pages}{3471--3477}.
\newblock


\bibitem[Wang et~al\mbox{.}(2021b)]%
        {wang2021neus}
\bibfield{author}{\bibinfo{person}{Peng Wang}, \bibinfo{person}{Lingjie Liu}, \bibinfo{person}{Yuan Liu}, \bibinfo{person}{Christian Theobalt}, \bibinfo{person}{Taku Komura}, {and} \bibinfo{person}{Wenping Wang}.} \bibinfo{year}{2021}\natexlab{b}.
\newblock \showarticletitle{NeuS: Learning Neural Implicit Surfaces by Volume Rendering for Multi-view Reconstruction}.
\newblock \bibinfo{journal}{\emph{Advances in Neural Information Processing Systems}}  \bibinfo{volume}{34} (\bibinfo{year}{2021}), \bibinfo{pages}{27171--27183}.
\newblock


\bibitem[Wang et~al\mbox{.}(2021a)]%
        {wang2021elevation}
\bibfield{author}{\bibinfo{person}{Yusheng Wang}, \bibinfo{person}{Yonghoon Ji}, \bibinfo{person}{Dingyu Liu}, \bibinfo{person}{Hiroshi Tsuchiya}, \bibinfo{person}{Atsushi Yamashita}, {and} \bibinfo{person}{Hajime Asama}.} \bibinfo{year}{2021}\natexlab{a}.
\newblock \showarticletitle{Elevation angle estimation in 2d acoustic images using pseudo front view}.
\newblock \bibinfo{journal}{\emph{IEEE Robotics and Automation Letters}} \bibinfo{volume}{6}, \bibinfo{number}{2} (\bibinfo{year}{2021}), \bibinfo{pages}{1535--1542}.
\newblock


\bibitem[Wang et~al\mbox{.}(2019a)]%
        {wang2019three}
\bibfield{author}{\bibinfo{person}{Yusheng Wang}, \bibinfo{person}{Yonghoon Ji}, \bibinfo{person}{Hanwool Woo}, \bibinfo{person}{Yusuke Tamura}, \bibinfo{person}{Atsushi Yamashita}, {and} \bibinfo{person}{Hajime Asama}.} \bibinfo{year}{2019}\natexlab{a}.
\newblock \showarticletitle{Three-dimensional underwater environment reconstruction with graph optimization using acoustic camera}. In \bibinfo{booktitle}{\emph{2019 IEEE/SICE International Symposium on System Integration (SII)}}. IEEE, \bibinfo{publisher}{Institute of Electrical and Electronics Engineers}, \bibinfo{address}{Paris, France}, \bibinfo{pages}{28--33}.
\newblock


\bibitem[Wang et~al\mbox{.}(2018)]%
        {wang20183d}
\bibfield{author}{\bibinfo{person}{Yusheng Wang}, \bibinfo{person}{Yonghoon Ji}, \bibinfo{person}{Hanwool Woo}, \bibinfo{person}{Yusuke Tamura}, \bibinfo{person}{Atsushi Yamashita}, {and} \bibinfo{person}{Asama Hajime}.} \bibinfo{year}{2018}\natexlab{}.
\newblock \showarticletitle{3D occupancy mapping framework based on acoustic camera in underwater environment}.
\newblock \bibinfo{journal}{\emph{IFAC-PapersOnLine}} \bibinfo{volume}{51}, \bibinfo{number}{22} (\bibinfo{year}{2018}), \bibinfo{pages}{324--330}.
\newblock


\bibitem[Westman et~al\mbox{.}(2020a)]%
        {westman2020theory}
\bibfield{author}{\bibinfo{person}{Eric Westman}, \bibinfo{person}{Ioannis Gkioulekas}, {and} \bibinfo{person}{Michael Kaess}.} \bibinfo{year}{2020}\natexlab{a}.
\newblock \showarticletitle{A theory of fermat paths for 3d imaging sonar reconstruction}. In \bibinfo{booktitle}{\emph{2020 IEEE/RSJ International Conference on Intelligent Robots and Systems (IROS)}}. IEEE, \bibinfo{publisher}{Institute of Electrical and Electronics Engineers}, \bibinfo{address}{Las Vegas, USA}, \bibinfo{pages}{5082--5088}.
\newblock


\bibitem[Westman et~al\mbox{.}(2020b)]%
        {westman2020volumetric}
\bibfield{author}{\bibinfo{person}{Eric Westman}, \bibinfo{person}{Ioannis Gkioulekas}, {and} \bibinfo{person}{Michael Kaess}.} \bibinfo{year}{2020}\natexlab{b}.
\newblock \showarticletitle{A volumetric albedo framework for 3D imaging sonar reconstruction}. In \bibinfo{booktitle}{\emph{2020 IEEE International Conference on Robotics and Automation (ICRA)}}. IEEE, \bibinfo{publisher}{Institute of Electrical and Electronics Engineers}, \bibinfo{address}{Virtual, Web}, \bibinfo{pages}{9645--9651}.
\newblock


\bibitem[Westman and Kaess(2019)]%
        {westman2019wide}
\bibfield{author}{\bibinfo{person}{Eric Westman} {and} \bibinfo{person}{Michael Kaess}.} \bibinfo{year}{2019}\natexlab{}.
\newblock \showarticletitle{Wide aperture imaging sonar reconstruction using generative models}. In \bibinfo{booktitle}{\emph{2019 IEEE/RSJ International Conference on Intelligent Robots and Systems (IROS)}}. IEEE, \bibinfo{publisher}{Institute of Electrical and Electronics Engineers}, \bibinfo{address}{Macau, China}, \bibinfo{pages}{8067--8074}.
\newblock


\bibitem[Yariv et~al\mbox{.}(2021)]%
        {yariv2021volume}
\bibfield{author}{\bibinfo{person}{Lior Yariv}, \bibinfo{person}{Jiatao Gu}, \bibinfo{person}{Yoni Kasten}, {and} \bibinfo{person}{Yaron Lipman}.} \bibinfo{year}{2021}\natexlab{}.
\newblock \showarticletitle{Volume {{Rendering}} of {{Neural Implicit Surfaces}}}. In \bibinfo{booktitle}{\emph{Advances in {{Neural Information Processing Systems}}}}, Vol.~\bibinfo{volume}{34}. \bibinfo{publisher}{Curran Associates, Inc.}, \bibinfo{address}{Virtual, Web}, \bibinfo{pages}{4805--4815}.
\newblock


\bibitem[Yariv et~al\mbox{.}(2023)]%
        {yariv2023bakedsdf}
\bibfield{author}{\bibinfo{person}{Lior Yariv}, \bibinfo{person}{Peter Hedman}, \bibinfo{person}{Christian Reiser}, \bibinfo{person}{Dor Verbin}, \bibinfo{person}{Pratul~P. Srinivasan}, \bibinfo{person}{Richard Szeliski}, \bibinfo{person}{Jonathan~T. Barron}, {and} \bibinfo{person}{Ben Mildenhall}.} \bibinfo{year}{2023}\natexlab{}.
\newblock \showarticletitle{{{BakedSDF}}: {{Meshing Neural SDFs}} for {{Real-Time View Synthesis}}}. In \bibinfo{booktitle}{\emph{{{ACM SIGGRAPH}} 2023 {{Conference Proceedings}}}} \emph{(\bibinfo{series}{{{SIGGRAPH}} '23})}. \bibinfo{publisher}{Association for Computing Machinery}, \bibinfo{address}{New York, NY, USA}, \bibinfo{pages}{1--9}.
\newblock
\showISBNx{9798400701597}
\urldef\tempurl%
\url{https://doi.org/10.1145/3588432.3591536}
\showDOI{\tempurl}


\bibitem[Yariv et~al\mbox{.}(2020)]%
        {yariv2020multiview}
\bibfield{author}{\bibinfo{person}{Lior Yariv}, \bibinfo{person}{Yoni Kasten}, \bibinfo{person}{Dror Moran}, \bibinfo{person}{Meirav Galun}, \bibinfo{person}{Matan Atzmon}, \bibinfo{person}{Basri Ronen}, {and} \bibinfo{person}{Yaron Lipman}.} \bibinfo{year}{2020}\natexlab{}.
\newblock \showarticletitle{Multiview neural surface reconstruction by disentangling geometry and appearance}.
\newblock \bibinfo{journal}{\emph{Advances in Neural Information Processing Systems}}  \bibinfo{volume}{33} (\bibinfo{year}{2020}), \bibinfo{pages}{2492--2502}.
\newblock


\bibitem[Zhu et~al\mbox{.}(2023)]%
        {10160388}
\bibfield{author}{\bibinfo{person}{Haidong Zhu}, \bibinfo{person}{Yuyin Sun}, \bibinfo{person}{Chi Liu}, \bibinfo{person}{Lu Xia}, \bibinfo{person}{Jiajia Luo}, \bibinfo{person}{Nan Qiao}, \bibinfo{person}{Ram Nevatia}, {and} \bibinfo{person}{Cheng-Hao Kuo}.} \bibinfo{year}{2023}\natexlab{}.
\newblock \showarticletitle{Multimodal {{Neural Radiance Field}}}. In \bibinfo{booktitle}{\emph{2023 {{IEEE International Conference}} on {{Robotics}} and {{Automation}} ({{ICRA}})}}. \bibinfo{publisher}{Institute of Electrical and Electronics Engineers}, \bibinfo{address}{London, England}, \bibinfo{pages}{9393--9399}.
\newblock
\urldef\tempurl%
\url{https://doi.org/10.1109/ICRA48891.2023.10160388}
\showDOI{\tempurl}


\end{thebibliography}
